\documentclass[letterpaper, 10 pt, conference]{ieeeconf}  

\IEEEoverridecommandlockouts                              

\usepackage{graphicx}
\usepackage[caption=false,font=normalsize,labelfont=sf,textfont=sf]{subfig}
\usepackage{stfloats}

\usepackage{amsmath}
\usepackage{amsfonts}

\usepackage[ruled]{algorithm2e}
\usepackage{hyperref}

\usepackage[dvipsnames]{xcolor}
\usepackage[deletedmarkup=sout, authormarkup=superscript]{changes}
\definechangesauthor[name={Todo}, color=Red]{TD}

\definechangesauthor[name={Movie_todo}, color=NavyBlue]{Movie}

\definechangesauthor[name={Bram vdb}, color=OliveGreen]{bv}

\definechangesauthor[name={Gaoyuan L}, color=Cyan]{gl}

\definechangesauthor[name={Naftali Slob}, color=Violet]{ns}

\definechangesauthor[name={Bas Boom}, color=Gray]{bb}

\definechangesauthor[name={Tom Turcksin}, color=NavyBlue]{tot}

\definechangesauthor[name={Yuri Durodie}, color=NavyBlue]{yd}

\usepackage{tikz,eso-pic}

\AddToShipoutPicture{%
\begin{tikzpicture}[remember picture, overlay]
      \node[minimum width=4in] at ([yshift=-1cm]current page.north)  {Preprint. The paper is published on IEEE Robotics \& Automation Magazine.};
    \end{tikzpicture}%

}

\title{\LARGE \bf
Automated Behavior Planning for Fruit Tree Pruning via Redundant Robot Manipulators: Addressing the Behavior Planning Challenge
}

\author{Gaoyuan Liu$^{1,4}$, Bas Boom$^{3}$, Naftali Slob$^{3}$, Yuri Durodié$^{1,2}$, Ann Nowé$^{4}$, Bram Vanderborght$^{1,2}$
\thanks{This work was funded by the \textit{Flemish} Government under the program \textit{Onderzoeksprogramma Artifici\H ele Intelligentie (AI) Vlaanderen}.}
\thanks{$^{1}$ Authors are with Brubotics, Vrije Universiteit Brussel, Brussels, Belgium. {\tt\small gaoyuan.liu@vub.be}}%
\thanks{$^{2}$ Authors are affiliated to imec, Belgium}%
\thanks{$^{3}$ Authors are affiliated to imec, the Netherlands}%
\thanks{$^{4}$ Ann Nowe is with the Artificial Intelligence (AI) Lab, Vrije Universiteit Brussel, Brussels, Belgium.}%
}

\begin{document}

\maketitle
\thispagestyle{empty}
\pagestyle{empty}

\begin{abstract}

Pruning is an essential agricultural practice for orchards. Proper pruning can promote healthier growth and optimize fruit production throughout the orchard's lifespan. Robot manipulators have been developed as an automated solution for this repetitive task, which typically requires seasonal labor with specialized skills. While previous research has primarily focused on the challenges of perception, the complexities of manipulation are often overlooked. These challenges involve planning and control in both joint and Cartesian spaces to guide the end-effector through intricate, obstructive branches.
Our work addresses the behavior planning challenge for a robotic pruning system, which entails a multi-level planning problem in environments with complex collisions. In this paper, we formulate the planning problem for a high-dimensional robotic arm in a pruning scenario, investigate the system's intrinsic redundancies, and propose a comprehensive pruning workflow that integrates perception, modeling, and holistic planning.
In our experiments, we demonstrate that more comprehensive planning methods can significantly enhance the performance of the robotic manipulator. Finally, we implement the proposed workflow on a real-world robot. As a result, this work complements previous efforts on robotic pruning and motivates future research and development in planning for pruning applications.
\end{abstract}

\section{Introduction}
\begin{figure}[h]
    \centering
    \subfloat[]{\includegraphics[width=0.23\textwidth]{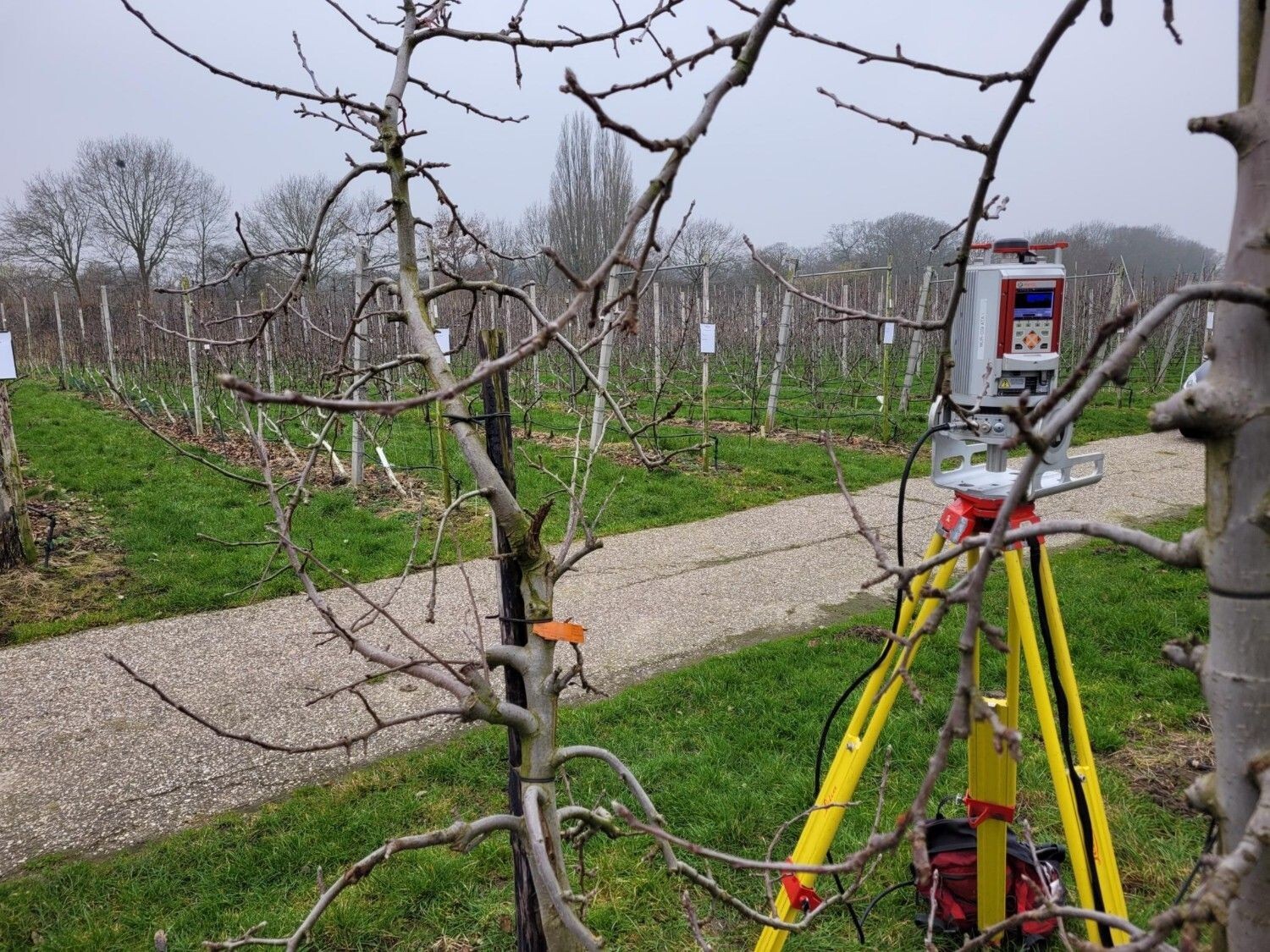}\label{fig:motivation_1}}
    \hfil
    \subfloat[]{\includegraphics[width=0.23\textwidth]{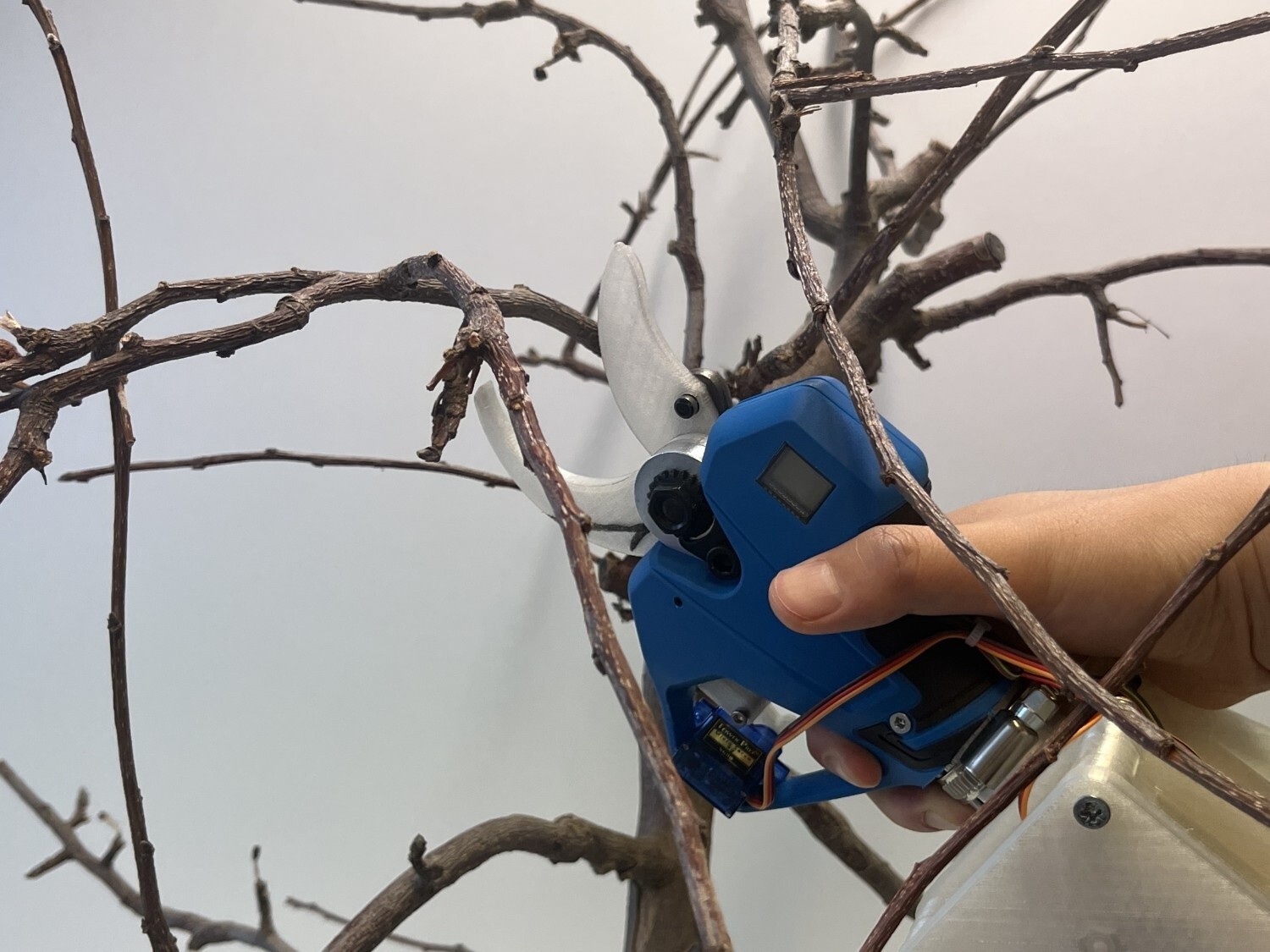}\label{fig:motivation_2}}
    \hfil
    \subfloat[]{\includegraphics[width=0.48\textwidth]{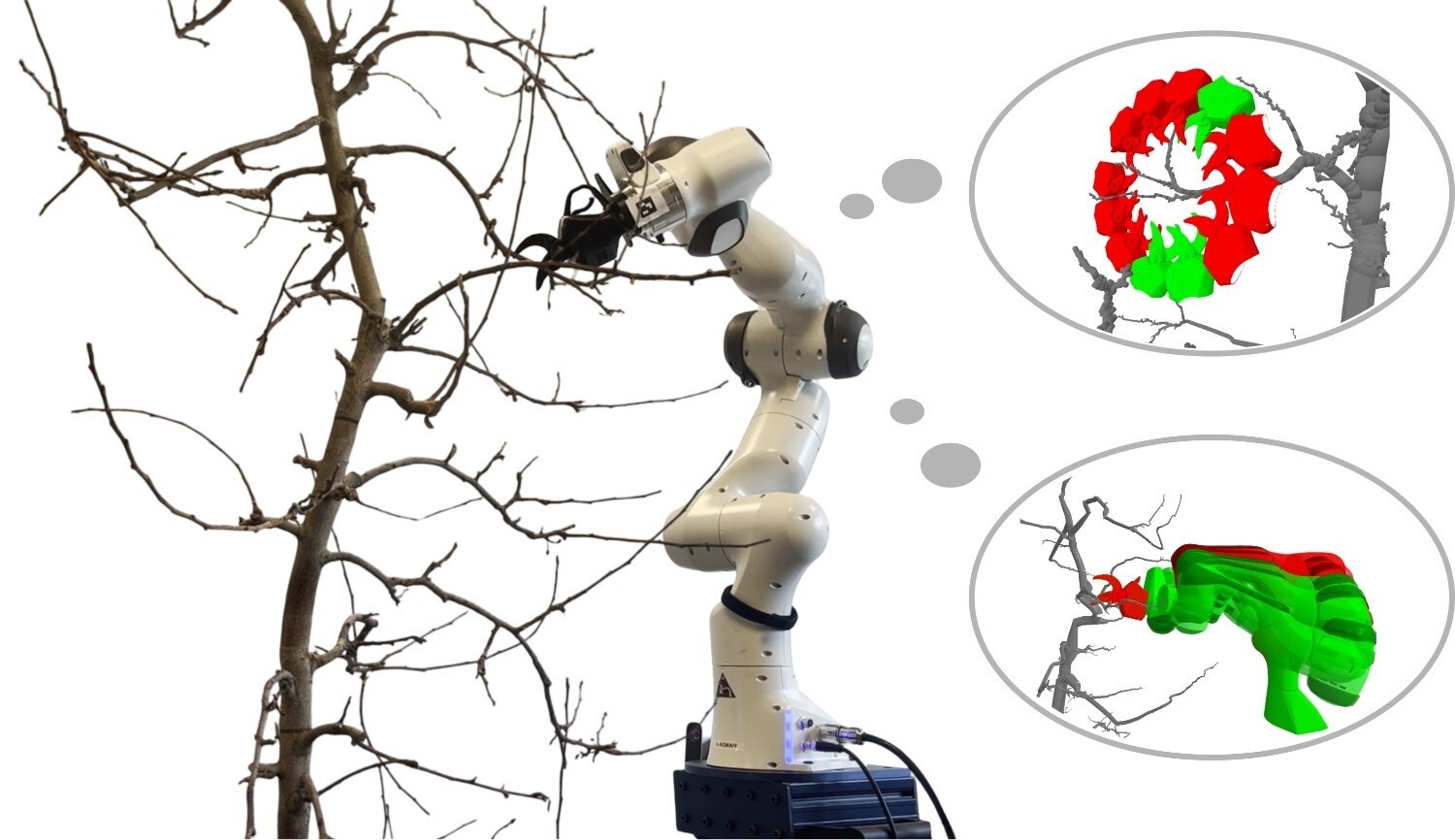}\label{fig:motivation_3}}
    \caption{The challenge of the pruning manipulation. (a) An apple orchard during the pruning season presents a unique challenge for robotic pruning. Despite the large-scale organization of the orchard, the complex arrangement of branches creates significant collision issues. (b) Human operators achieve a high success rate by skillfully selecting approaching poses that comprehensively consider collisions and accessibility. (c) In order to tackle the intricate collision during pruning, the robot needs to explore the redundancy in both Cartesian and joint space. The green color indicates the valid solutions while the red color indicates collisions.}
    \label{fig:intro}
\end{figure}

Pruning is a crucial procedure in cultivation, as it involves removing dead and unproductive branches and stubs, creating space for new growth. However, manual pruning requires a considerable number of seasonal laborers and involves an extensive workload. The shortage of skilled pruning workers and the inherent hazards of the physical task increase the costs in the fruit industry.
Automatic pruning has been introduced to tackle these challenges \cite{tinoco2021review}. One prevalent method in automated pruning mechanization is non-selective pruning, commonly known as hedging. This approach decides which branches to trim solely based on the distance between the tree and the trimming bar \cite{ferree1993apple}. However, this crude method may lead to damage, as productive outlying branches may be inadvertently cut off, thereby reducing the cumulative yield per tree.
Selective pruning with a high-dimensional robotic manipulator offers an agile alternative for the pruning task, requiring a robot with the flexibility to avoid branches while still effectively reaching the desired goal position \cite{tinoco2021review}. Robotic pruning demands that the robot possess various skills, including branch detection, cutting sequence decision-making, and collision-free, precise manipulation. These abilities are crucial for ensuring efficient and accurate pruning operations. Previous efforts have primarily focused on tree detection using computer vision technologies, such as image segmentation \cite{9811628}, depth image mapping \cite{9197551}, and reconstruction \cite{you2024realtimehardwareagnosticframework}.
However, the pruning manipulation problem has often been neglected or oversimplified, as the structure of target plants is typically considered relatively simple and uniform. For example, previous studies have focused on grapevines \cite{botterill2017robot} and cherry trees \cite{you2023semiautonomous}.
For an apple orchard, as shown in Figure \ref{fig:motivation_1}, the tree structures are more spatial than planar, meaning their branches grow in arbitrary directions. Pruning trees with such structures requires navigating a complex collision space \cite{zahid2020collision}, as human operators skillfully select approach angles and motions while accounting for occlusions, as illustrated in Figure \ref{fig:motivation_2}. The complexity of the tree structure makes basic motion planning inefficient due to a low success ratio. Intuitively, this low success ratio arises from the narrow solution space. For example, as shown in Figure \ref{fig:motivation_3}, the green color indicates collision-free cutting poses and configurations, while the red color indicates colliding ones. Thus, the solution space lies in the narrow configuration space. To improve the planner's capability and the success ratio, we explore the redundancy inherent in the robotic pruning task. We argue that, in pruning, redundancies exist in both Cartesian and joint space. The redundancy in Cartesian space arises because a cutting location can be approached from arbitrary angles surrounding the branch. The redundancy in joint space stems from the robot's extra degrees of freedom (DoF). These redundancies expand the search space of the planning problem and, as demonstrated in our experiments, improve the success ratio.

In this paper, we investigate behavior design for pruning with a robotic manipulator. We design a two-stage pruning motion and a holistic planning strategy that accounts for constraints such as obstacles, singularities, and joint limits. This effort is intended to complement previous work, which has primarily focused on perception. On top of that, a robotic pruning system with a software workflow is developed. The system requires an industrial manipulator and a depth camera. The workflow includes perception, modeling, planning, and control procedures. With the point cloud data gathered from the depth camera, an \textit{AdTree} \cite{rs11182074} model containing detailed geometrical information is constructed. AdTree is one of the benchmarking point cloud reconstruction methods for trees. It is extensively used in tree-related applications, such as biomass estimation \cite{fan2020adqsm}, ecological modeling \cite{keerthinathan2025modelling}, and, alongside its primary focus on detailed tree reconstruction \cite{gao2025extraction}. The geometric model allows the robot to explore the aforementioned redundancies and expand the search space for the manipulation. We verify the method on a tree dataset generated from a real orchard and finally implement and demonstrate the workflow on a physical robot with a real tree sample.

The paper is organized as follows. Section \ref{sec:related_work} presents an overview of the related work. Section \ref{sec:methodology} details the methods and concepts in our framework. Section \ref{sec:experiments} demonstrates and analyzes the experiment results. Section \ref{sec:conclusion} concludes the paper.

\section{Related Work}\label{sec:related_work}

To realize an automated robotic pruning system, two primary challenges must be addressed: perception (sensing and modeling) and manipulation (planning and control) \cite{you2023semiautonomous}. Previous research predominantly focuses on the former \cite{you2022optical, borrenpohl2023automated, you2024realtimehardwareagnosticframework, 10384649, rs11182074, LI2023108149}, whereas investigations into planning and control for pruning tasks remain limited. 
Some previous works have focused on specific hardware designs for pruning \cite{chen2022path, zahid2020development}. Zahid et al. developed a 6-DoF apple tree pruning manipulator with three prismatic joints and three revolute joints \cite{zahid2020development}. They concluded that future studies should focus on collision-free motion planning, which is the focus of this work. Although customized systems can achieve a better working range, the design and verification cycle of hardware development is costly. Additionally, customized robots are exclusively dedicated to a single task, limiting their market potential. In this paper, we aim to leverage market-available devices and seek a solution from a software perspective. Commercial manipulators, such as the UR series, have been used in pruning tasks. A notable example is the grapevine pruning robotic platform \cite{botterill2017robot, silwal2021bumblebee, molaei2022kinematic}. This platform covers the target tree structure to ensure uniform lighting conditions \cite{botterill2017robot}. The study extends previous work by integrating more advanced control strategies, such as reinforcement learning \cite{9562075, 9811628}. However, the structure of the grapevine is relatively simple, and the robot can collide with the canes due to their softness \cite{silwal2021bumblebee}. In contrast, when the target plant is more rigid or fragile, collisions must be carefully considered.
You et al. used an octomap to represent collisions and a sampling-based motion planner to generate a collision-free path \cite{9197551}. Since the tree structure is approximately planar, the approach poses are pre-discretized as a 2D matrix. However, such strategies are insufficient when the tree structure is more spatial. In \cite{9811628}, the pre-defined approach poses are replaced by an RL policy. Their method requires the robot's end-effector to be positioned 15-20 cm from the target branch, ensuring the Cartesian path is collision-free. Similar to \cite{zahid2020collision}, we eliminate the aforementioned requirements by building a geometric collision model of the tree and a motion planner. Instead of relying solely on a motion planner or unrealistic virtual tree models, we inherit the two-stage strategy from \cite{9811628} and use real tree models.
With the perception and manipulation subsystems, previous works have already achieved complete robotic pruning systems for orchards \cite{you2023semiautonomous, YOU2022106622}. Instead of a discretized pose matrix, a pre-defined waypoint trajectory is provided to avoid obstacles in \cite{you2023semiautonomous}. The trajectory improved the adaptivity of the previous strategy by adjusting according to the distribution of the branches. Most previous applications are limited to (nearly) planar trees such as grapevine and cherry trees. Our work extends the previous work in a more general spatial tree structure, i.e., a spindle structure, where the branches are extending from arbitrary directions.

\section{Methodology}\label{sec:methodology}

\begin{figure*}
    \centering
    \vspace{2 mm}
    \includegraphics[width=0.98\textwidth]{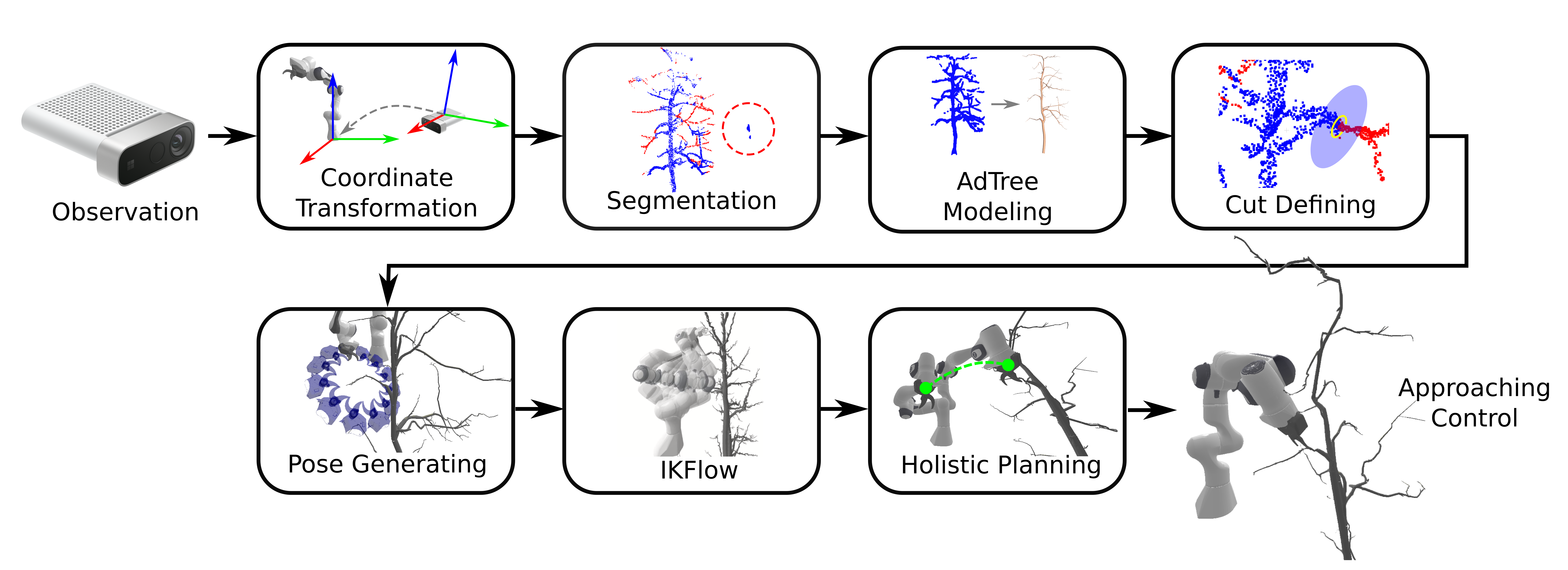}
    \caption{The workflow of our system consists of multiple modules. The system's input is the point cloud data captured by a depth camera. The following procedures will be conducted: (1) Coordinate Transformation: The observation data needs to be transformed from the camera frame to the robot's frame to align perception and robotic behavior. (2) Segmentation: The observed data tend to contain environmental noise such as ground and outliers. (3) AdTree Modeling: Generate an AdTree from the point cloud data. (4) Cut Defining: With labeled point cloud data, we can define cutting locations on the AdTree. (5) Pose Generating: For each cutting location, the planner will generate multiple potential cutting poses to approach the cutting position and choose one based on accessibility. (6) IKFlow: A learning-based IK solver, it can provide diverse IK solutions for one end-effector pose. (7) Holistic Planner: A planner that comprehensively plans the end-effector pose, joint configuration, and trajectory. }
    \label{fig:workflow}
\end{figure*}

\begin{figure}
    \centering
    \vspace{2 mm}
    \subfloat[]{\includegraphics[width=0.4\textwidth]{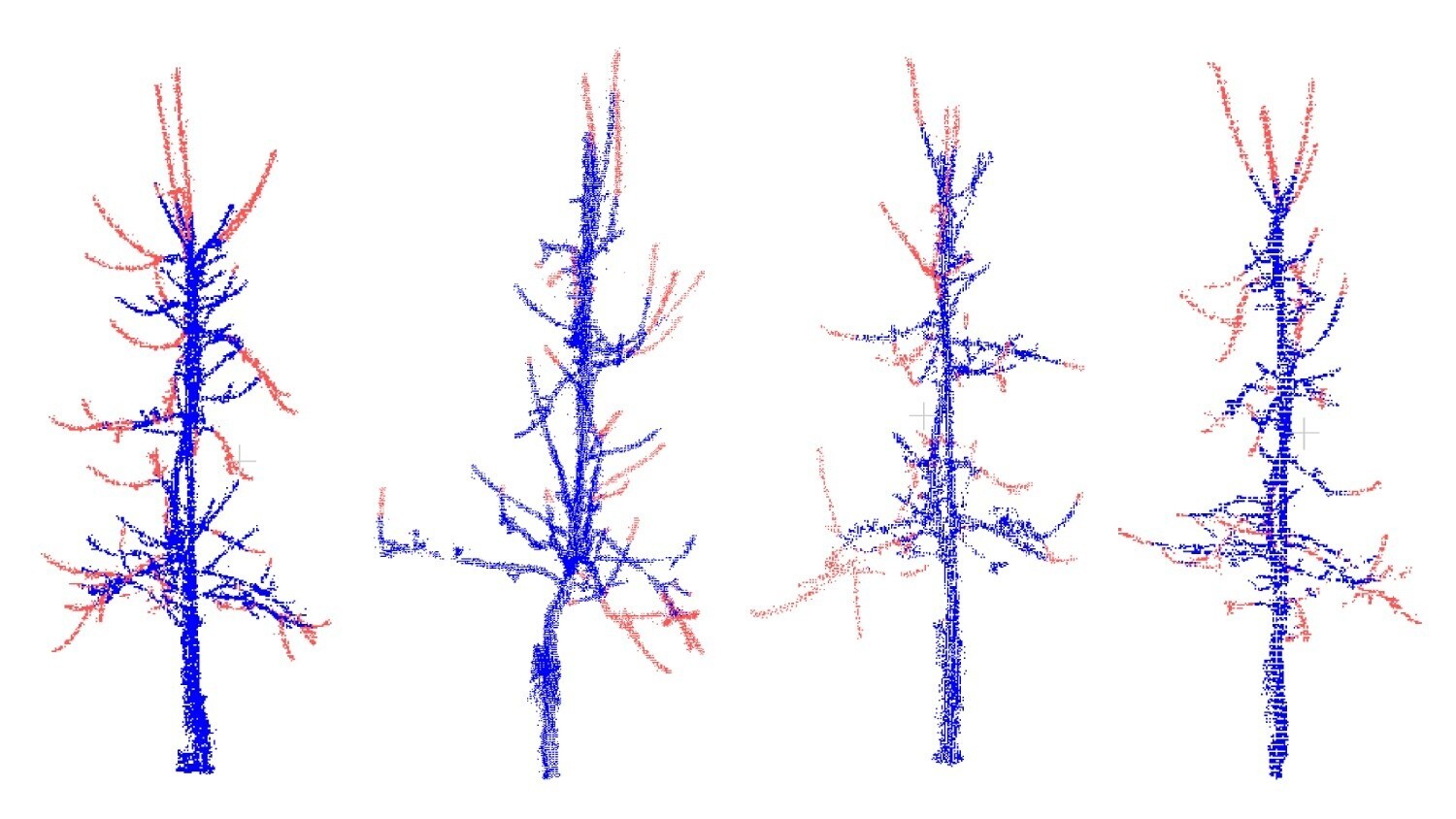}\label{fig:annotated_pc}}
    \hfil
    \subfloat[]{\includegraphics[width=0.36\textwidth]{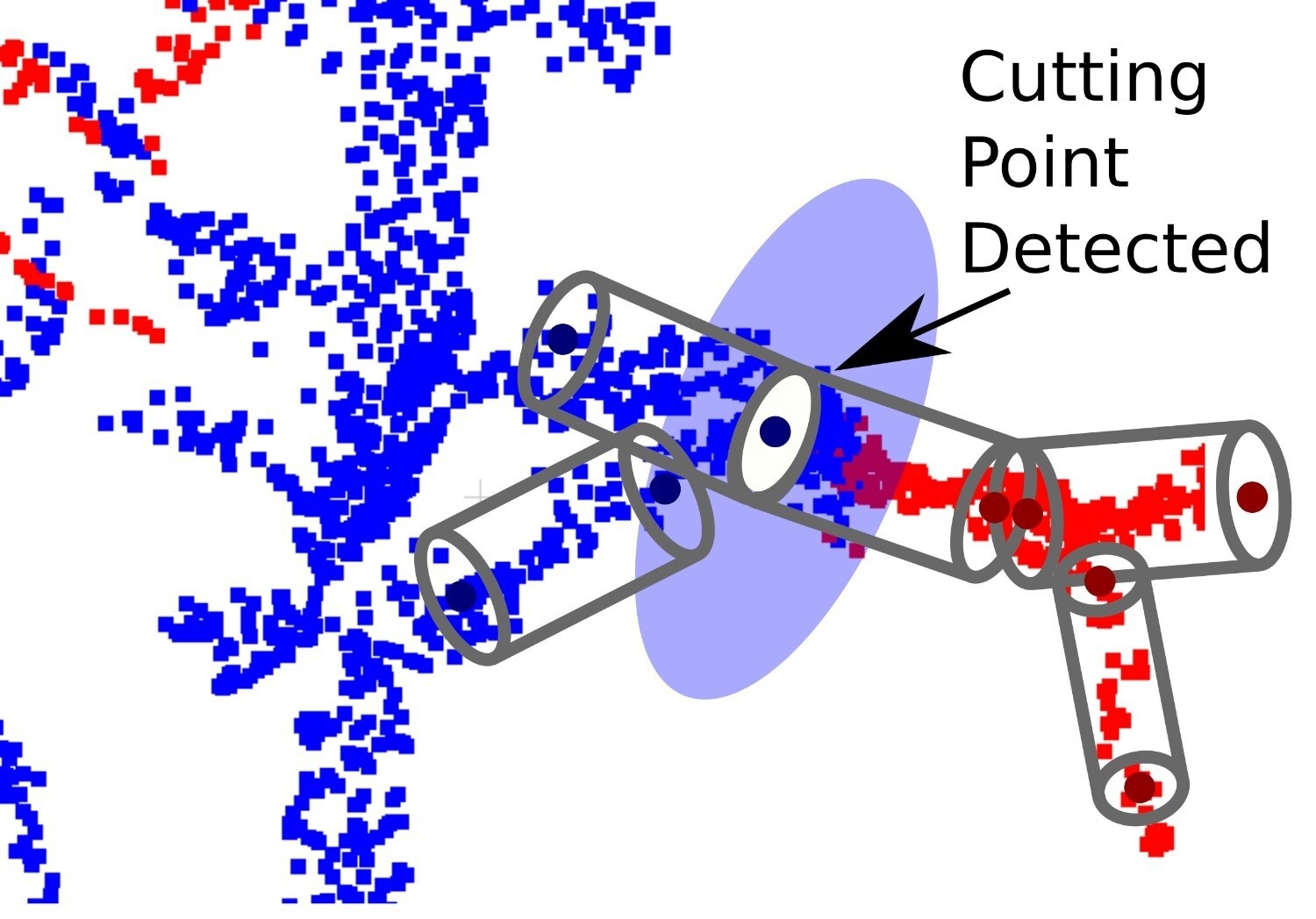}\label{fig:cutting_method}}
    \caption{(a) Labeled point cloud input: Our method takes a labeled point cloud as input. In the example shown, red points represent branches to be removed, while blue points represent parts to be kept. (b) To identify cutting locations in the labeled point cloud on the AdTree, we first assign labels to the AdTree's vertices. A cutting location is detected on an AdTree vertex when a consecutive point has a different label. In the figure, the gray cylinders represent sections of the AdTree, and the large dots inside the cylinders represent vertices, with their colors indicating the labels.}
\end{figure}

Pruning planning presents a complex domain characterized by continuous spaces and intricate constraints. Successfully addressing this planning problem entails creating models for the intricate collision environments and implementing a planning system capable of navigating in highly constrained environments. In this section, we propose a workflow for robotic pruning, as shown in Figure \ref{fig:workflow}. This pipeline takes as input a labeled point cloud of a tree, where the points to be pruned by the robot are marked with a specific label, i.e., $\mathcal{L}_{\bullet} = \{0,1\}$. As shown in Figure \ref{fig:annotated_pc}, $\mathcal{L}_{\bullet}$ is a list of labels per point in the point cloud having either value $0$ (points in blue indicate to remain) or $1$ (points in red indicate to remove). Section \ref{sec:tree_dataset} discusses several ways to obtain such annotated point clouds with associated labels. Although the capture of the point cloud is not the primary focus, our system currently supports both LiDAR data capture in a real orchard and Time of Flight (ToF) data captured in our laboratory. Given the point cloud, the pipeline generates a topological structure with branch information using the AdTree method \cite{rs11182074}. This topological structure is then transferred to a geometric model within the robotic simulation environment. Such a model enables the command to be formulated with geometrical information, allowing the downstream planning routine to exploit the redundancy in the Cartesian space.
Moreover, the topological information enables the concave geometry of the entire tree to be decomposed into convex primitive shapes, such as cylinders, which can be used for collision detection during planning.
The following sections discuss sub-modules in Figure \ref{fig:workflow}. The coordinate transformation and segmentation steps use basic methods such as Euclidean clustering and axis-aligned bounding box. An introductory video, including experiments, can be found on the website: \url{https://gaoyuan-liu.github.io/robotic-pruning/}. 

\subsection{Tree Modeling and Cut Defining}\label{sec:tmacld}

\begin{figure}[h]
    \centering
    \vspace{2 mm}
    \subfloat[]{\includegraphics[height=0.22\textwidth]{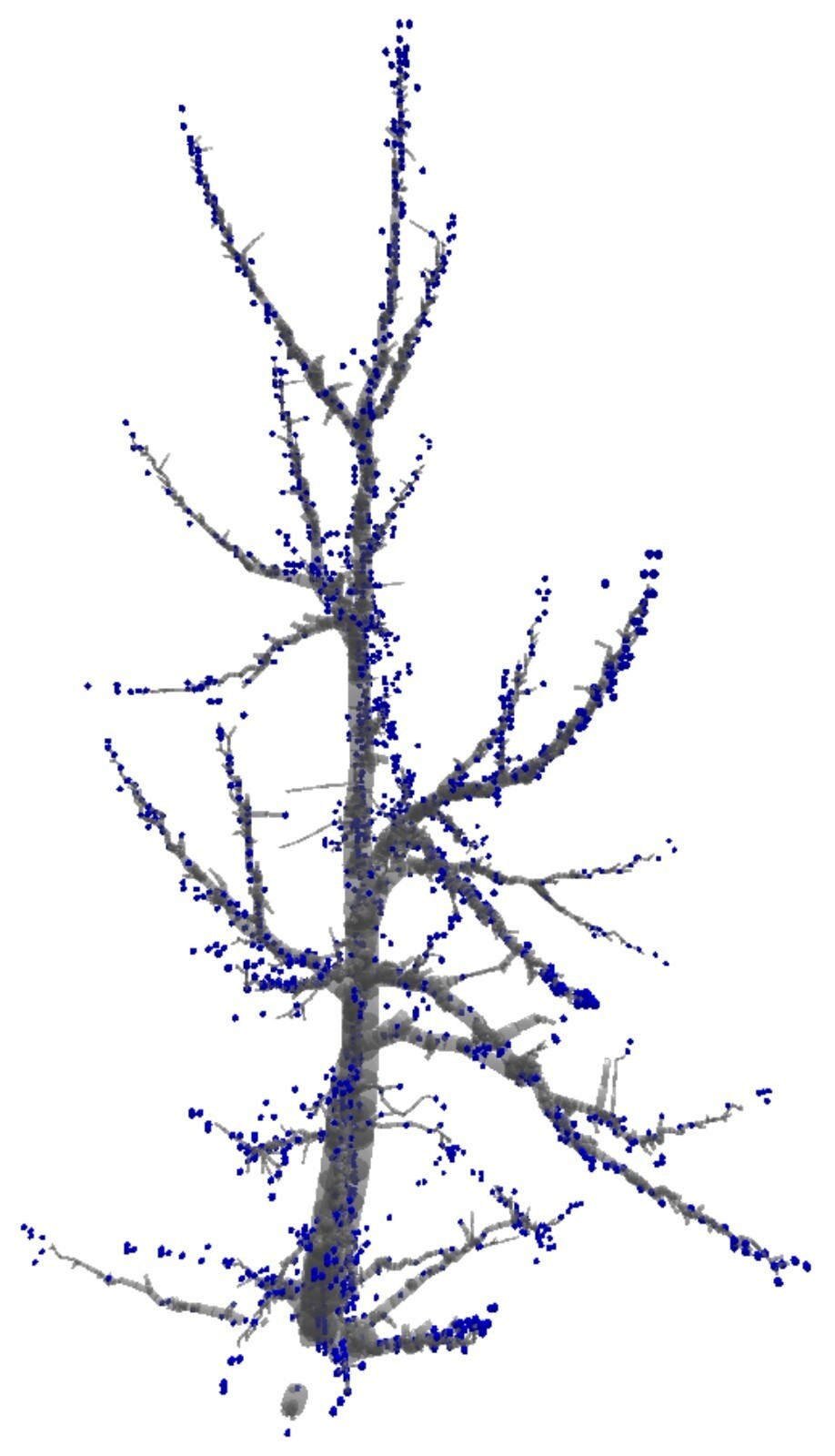}\label{fig:pc2adtree_1}}
    \hfil
    \subfloat[]{\includegraphics[height=0.22\textwidth]{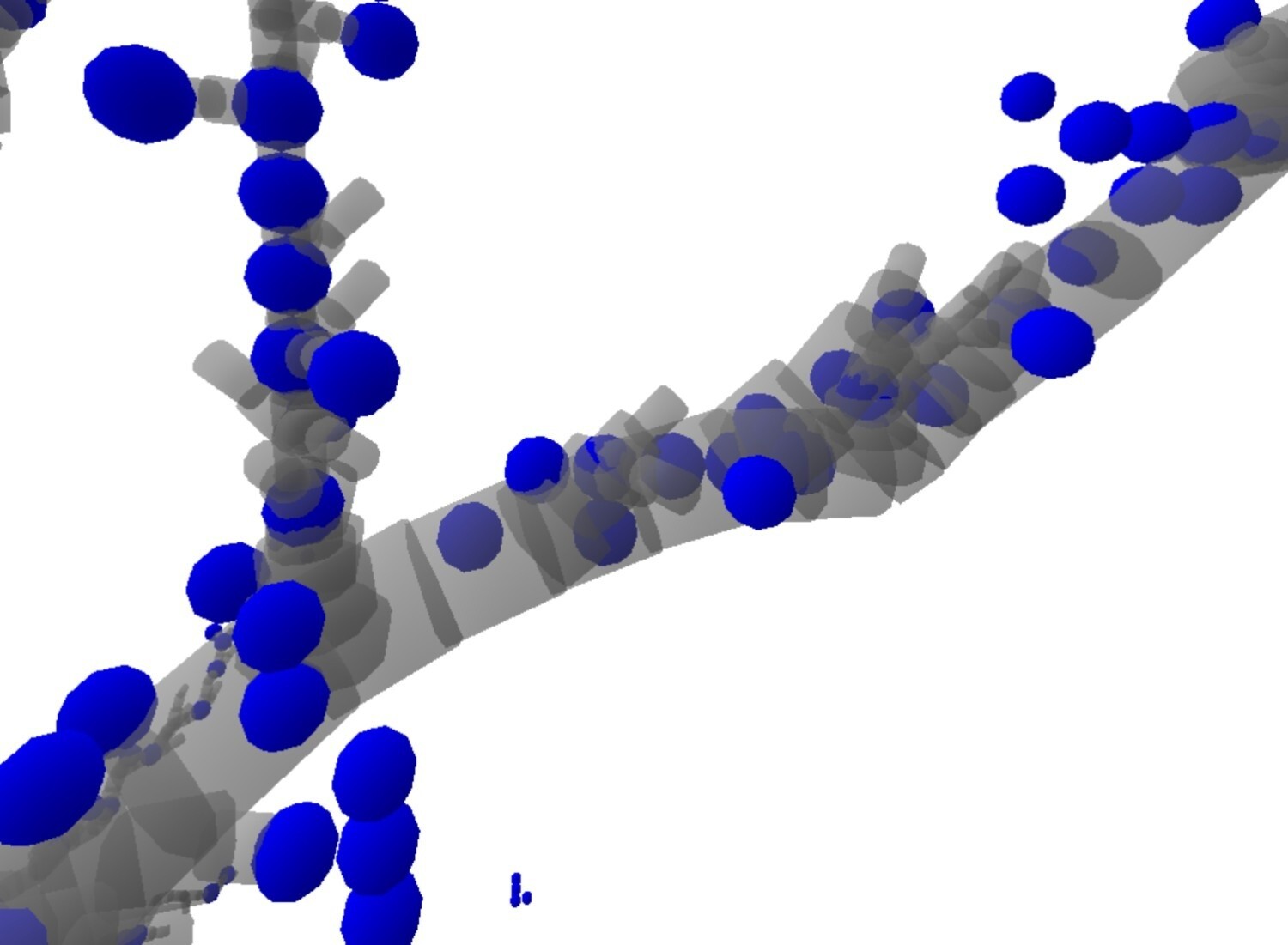}\label{fig:pc2adtree_2}} 
    \caption{The visualization illustrates the AdTree. AdTree abstracts a tree model from the point cloud. Its topological structure is represented as a geometric tree in the simulation. In the figure, cylinders are generated based on vertex, edge, and thickness data, visually representing different parts of the tree. The blue dots represent the point cloud. }
    \label{fig:pc2adtree}
\end{figure}

Given a point cloud of a tree with labels $\mathcal{L}{\bullet}$, the point cloud is converted into a topological structure with branch information using AdTree. As shown in Figure \ref{fig:pc2adtree}, AdTree is a method that accurately and automatically reconstructs detailed 3D tree models from point clouds. The output of AdTree is a graph with both vertices $V$ and edges $E$, where the edge information also includes an estimation of the branch diameter. With this estimation, each edge can be represented as a cylinder segment. Therefore, the geometric model of the tree can be generated by connecting all the cylinder segments. To utilize the graph, a k-d tree is employed to search the point cloud for the closest points matching the vertices in the graph, in order to obtain the labels per vertex ($\mathcal{L}{\boldsymbol{v}}$). Subsequently, Algorithm \ref{alg:cut_location} is used to identify a list of cutting commands. We define each cutting command as a \textit{cut}, i.e., $\boldsymbol{c} = (\boldsymbol{t}_c, \overrightarrow{\boldsymbol{v}}_c)$, which consists of a cutting position $\boldsymbol{t}_c$ and the norm of the cutting section surface $\overrightarrow{\boldsymbol{v}}_c$. In Algorithm \ref{alg:cut_location}, we recursively walk over the tree graph defined by AdTree, obtain the cut position, and compute the cut vector based on the label change. Figure \ref{fig:cutting_method} illustrates how the cutting location is defined with labeled point clouds.

\begin{algorithm}
\caption{Cut Generation}\label{alg:cut_location}

\textbf{Given} the topological structure of tree with vertices $\mathcal{V}$ and edges $\mathcal{E}$, together with a list of labels per vertex $\mathcal{L}_{\boldsymbol{v}}$. \\

\textbf{Given} the set of vertices in memory $\mathcal{S} = \{v_{root}\}$\\

\textbf{Given} the output set of cut points $\mathcal{C} = \emptyset$ \\

\While{$\mathcal{S} \neq \emptyset$}{
    \textcolor{blue}{\# Get last element from $\mathcal{S}$}\\
    $\boldsymbol{x} := \mathcal{S}_{\text{last}}$\\
    
    $\mathcal{S} := \mathcal{S} \setminus \{\boldsymbol{x}\}$\\

    \textcolor{blue}{\# Get the second elements for all edges where the first element is equal to vertex $\boldsymbol{x}$}\\
    $\mathcal{Y} := \{ \boldsymbol{e} \in \mathcal{E} : \boldsymbol{e}_0 = \boldsymbol{x} \}_{1}$\\

    $\mathcal{S} := \mathcal{S} \cup \mathcal{Y}$\\

    \For{$\boldsymbol{y} \in \mathcal{Y}$}{
        \If{$\mathcal{L}_{\boldsymbol{x}} = 0 \wedge \mathcal{L}_{\boldsymbol{y}} = 1$}{
            $\boldsymbol{t}_{c} := \boldsymbol{y}$\\
            $\overrightarrow{\boldsymbol{v}}_{c} := \boldsymbol{y} - \boldsymbol{x}$\\
            $\boldsymbol{c} := (\boldsymbol{t}_{c}, \overrightarrow{\boldsymbol{v}}_{c})$\\
            $\mathcal{C} := \mathcal{C} \cup \{ \boldsymbol{c} \}$\\
        }
    }
}
\end{algorithm}


\subsection{Robot Behavior Planning}
\begin{figure}
    \centering
    \includegraphics[width=0.45\textwidth]{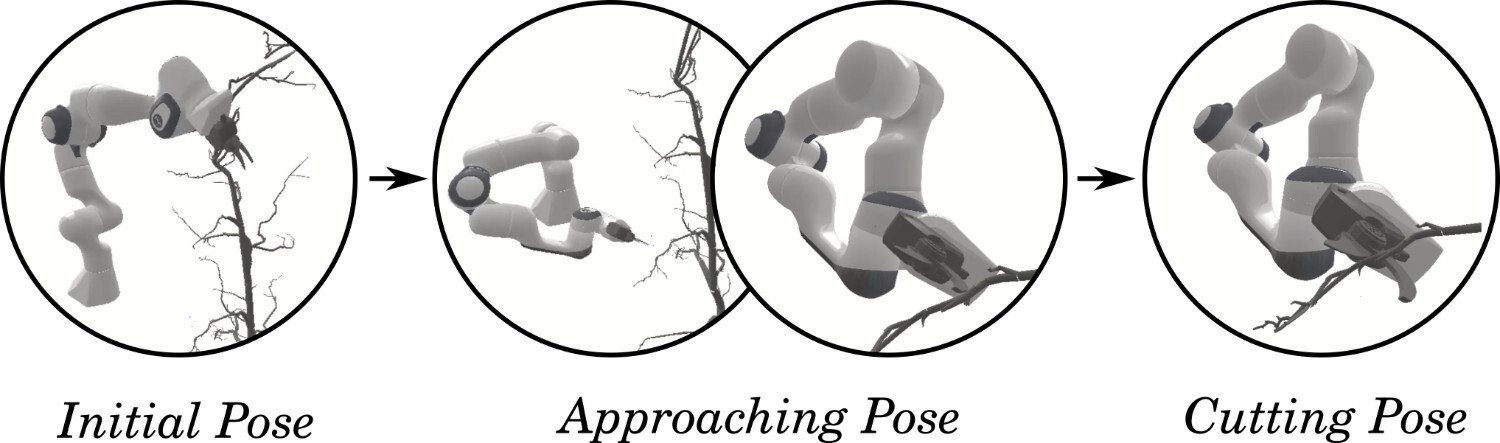}
    \caption{In the first stage, the robot moves from its initial pose to an approaching pose using a collision-free motion plan. In the second stage, the robot advances from the approaching pose to the cutting pose, positioning the target branch between the shears before making the cut. The approaching pose is shown from both a wide view and a close-up perspective. }
    \label{fig:planning_abbr}
\end{figure}

The planning problem for a spatial pruning task inherently involves multiple levels of planning, each offering redundancies that enable successful task execution. Theoretically, for any given cutting command on a branch, there are infinitely many possible poses for the robot's end-effector to approach it. Furthermore, for each approaching pose, there are infinite inverse-kinematics (IK) solutions and joint-space paths due to the robot's redundancy. While this abundance of possible robot states increases the flexibility of the solution space, the pruning task is performed in a highly constrained environment. For instance, the planning environment includes the complex collision distribution of the branches.
In this section, we design a robot behavior planning framework customized for the pruning task. As shown in Figure \ref{fig:planning_abbr}, we decompose the pruning motion into two stages: the moving stage and the approaching stage. In the moving stage, the robot moves to an \textit{approaching} \textit{pose} from its initial pose while avoiding obstacles; in the approaching stage, the robot goes from the \textit{approaching pose} to the \textit{cutting pose} while adjusting the end-effector pose to improve accuracy. Planning this two-stage motion requires a holistic planner that includes the following considerations: (1) The approaching pose and the cutting pose need to be collision-free and IK-solvable; (2) The IK solution for the approaching pose should provide ample margins from the limits, ensuring that the second stage can be executed smoothly without violating the constraints such as collisions and joint limits. 

\subsubsection{Pose Generating}
\begin{figure}
    \centering
    \subfloat[]{\includegraphics[width=0.25\textwidth]{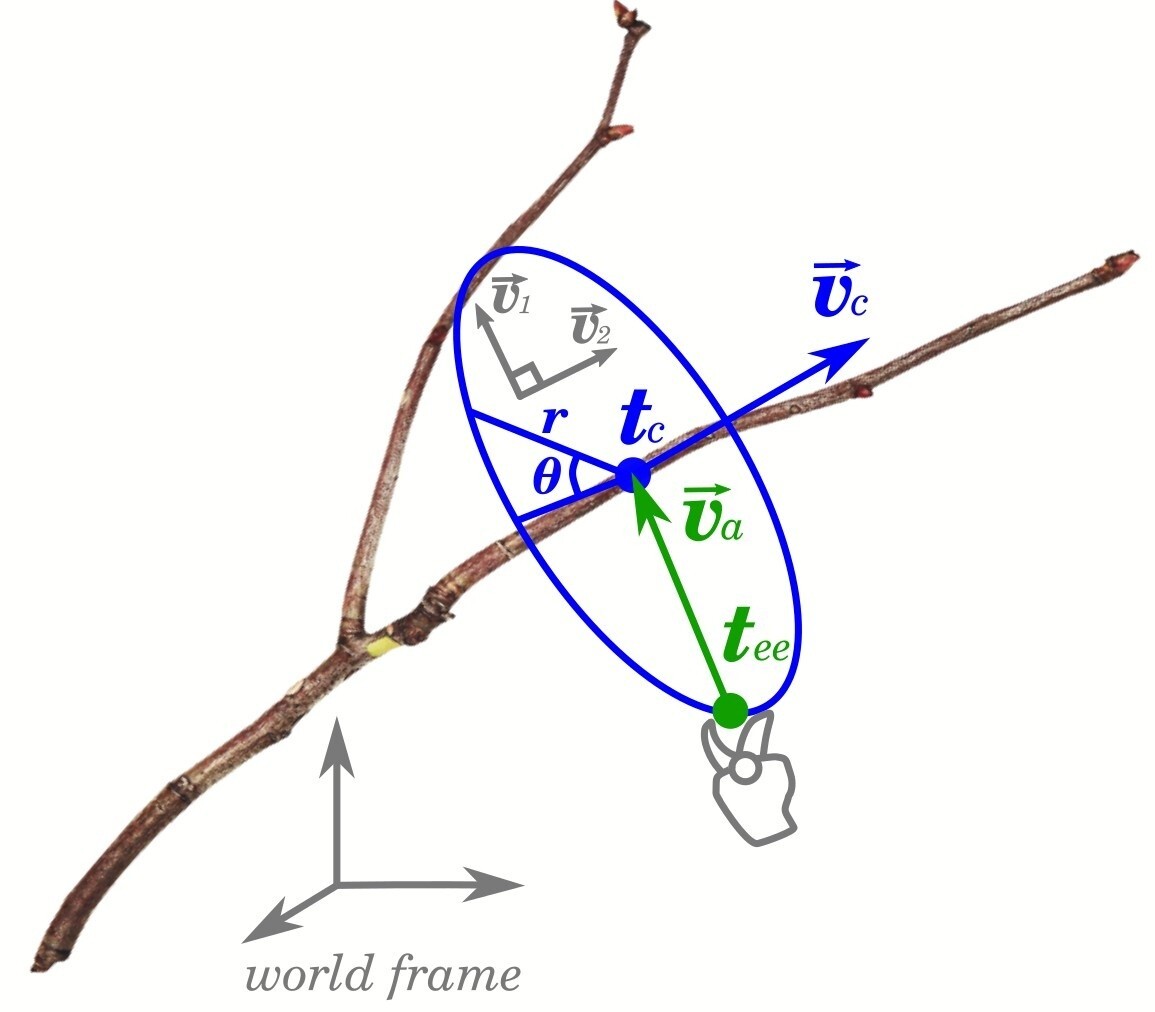}\label{fig:pose_sampling_1}}
    \hfil
    \subfloat[]{\includegraphics[width=0.2\textwidth]{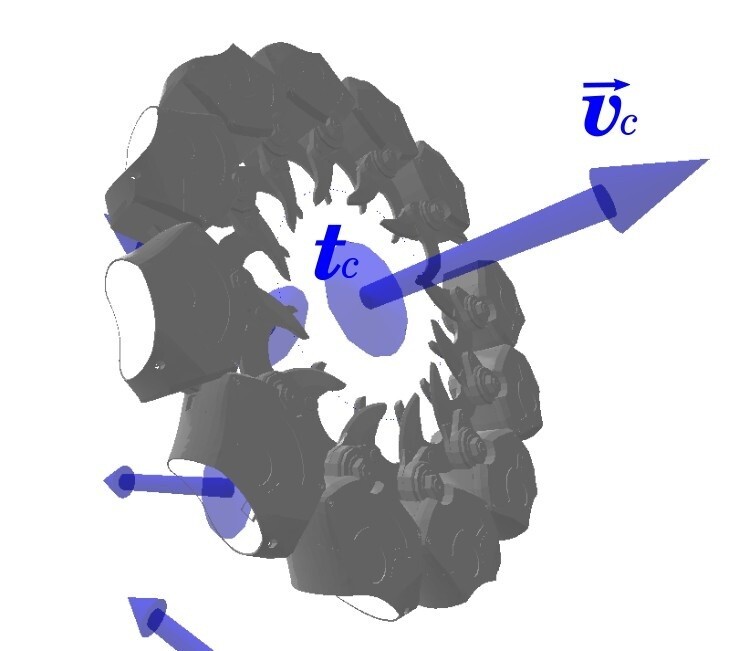}\label{fig:pose_sampling_2}}
    \caption{Pose generating. (a) Vectors, positions, and other variables that are used in the pose-generating process. The approaching and cutting positions lay in circles around the cutting vector. The orientations are calculated based on their spatial relations. (b) The generated approaching poses for the end-effector. The corresponding cutting command, $\boldsymbol{c}=(\boldsymbol{t}_c, \overrightarrow{\boldsymbol{v}}_c)$, is illustrated by the blue dot and arrow. }
    \label{fig:pose_sampling_method}
\end{figure}

Due to the uniform nature of a pruning task, we can approach the cutting location from any angle, allowing for greater flexibility and adaptability in the cutting process. We generate the potential approaching and cutting poses of the end-effector by calculating poses on the surrounding circle of the cutting position on the branch. As discussed in Sector \ref{sec:tmacld}, we assume that the branch is approximated as a cylinder near the cutting location on the tree. Therefore, the candidate approaching and cutting poses can be defined on a circle around it (see Figure \ref{fig:pose_sampling_method}).
We define the end-effector \textit{pose}, i.e., $\boldsymbol{p}_{ee} = (\boldsymbol{t}_{ee}, \boldsymbol{o}_{ee})$, is comprised of a position $\boldsymbol{t}_{ee}$ and a orientation $\boldsymbol{o}_{ee}$. 
By default, the orientation is formulated as a quaternion, i.e., $\boldsymbol{o}\in \mathbb{R}^4$. The position $\boldsymbol{t}$ is a tuple used to describe a spatial point without orientation. The approaching poses and the cutting poses are denoted as $\{ \boldsymbol{p}_{ee}^{a} \}$ and $\{ \boldsymbol{p}_{ee}^{c} \}$ respectively. 

The end-effector position can be formalized as follows:
\begin{equation}
\boldsymbol{t}_{ee} = \boldsymbol{t}_c+r\frac{\cos \theta}{||\overrightarrow{\boldsymbol{v}}_1||} \overrightarrow{\boldsymbol{v}}_1+r\frac{\sin \theta}{||\overrightarrow{\boldsymbol{v}}_2||} \overrightarrow{\boldsymbol{v}}_2
\end{equation}
Where $r$ is the distance from the end-effect position to the central axis of the branch cylinder, i.e., $r = ||\boldsymbol{t}_c - \boldsymbol{t}_{ee}||$, $\theta$ is the approaching angle that defines the position on the circle, $\overrightarrow{\boldsymbol{v}}_1$ and $\overrightarrow{\boldsymbol{v}}_2$ are two arbitrary orthogonal vectors on the cutting section surface, that is, $\overrightarrow{\boldsymbol{v}}_c$, $\overrightarrow{\boldsymbol{v}}_1$ and $\overrightarrow{\boldsymbol{v}}_2$ are orthogonal. Given $\overrightarrow{\boldsymbol{v}}_c = [v_{x}, v_{y}, v_{z}]$, that is not $v_{x}=v_{y}=0$, we can find:
\begin{align*} 
\overrightarrow{\boldsymbol{v}}_1 &= [v_{y}, -v_{x}, 0] ^ \top\\
\overrightarrow{\boldsymbol{v}}_2 &= [v_{x}v_{z}, v_{y}v_{z}, -v_{x}^{2}-v_{y}^{2}] ^ \top
\end{align*}
By choosing different $\theta$ and $r$, the end-effector position can be calculated.

After the end-effector position $\boldsymbol{t}_{ee}$ is settled, the orientation of the end-effector can also be determined by aligning the frames. We assume the desired cutting orientation is perpendicular to the cutting normal vector $\overrightarrow{\boldsymbol{v}}_c$, and the shear is facing the cutting position $\boldsymbol{t}_c$. Therefore, the orientation of the end-effector can be abstracted as the rotation matrix:
\begin{equation}
    \boldsymbol{R}_{ee} = 
    (\hat{\overrightarrow{\boldsymbol{v}}}_{c},  
    \hat{\overrightarrow{\boldsymbol{v}}}_{c \times a},
    \hat{\overrightarrow{\boldsymbol{v}}}_{a}
    )
\end{equation}

Where $\hat{\overrightarrow{\boldsymbol{v}}}_{c} = \overrightarrow{\boldsymbol{v}}_{c}/|\overrightarrow{\boldsymbol{v}}_{c}|$ is the normalized $\overrightarrow{\boldsymbol{v}}_{c}$, $\hat{\overrightarrow{\boldsymbol{v}}}_{a} = \overrightarrow{(\boldsymbol{t}_{ee}, \boldsymbol{t}_c)} / |\overrightarrow{(\boldsymbol{t}_{ee}, \boldsymbol{t}_c)}|$ is the unit vector from point $\boldsymbol{t}_{ee}$ to the direction of point $\boldsymbol{t}_c$, and $\hat{\overrightarrow{\boldsymbol{v}}}_{c \times a}$ is their vector product, i.e., $\hat{\overrightarrow{\boldsymbol{v}}}_{c \times a} = (\overrightarrow{\boldsymbol{v}}_{c} \times \overrightarrow{\boldsymbol{v}}_{a})/|\overrightarrow{\boldsymbol{v}}_{c} \times \overrightarrow{\boldsymbol{v}}_{a}|$. Finally, the orientation of the end-effector can be formatted as quaternion:

\begin{equation}
    \boldsymbol{o}_{ee} = \textbf{quat}(\boldsymbol{R}_{ee})
\end{equation}
Where $\textbf{quat}(\cdot)$ represents the calculator who transfers the rotation matrix to the quaternion. The advantage of this approach of pose generating is the distance $r$ and angle $\theta$ can be given explicitly. We therefore formulate the pose-generating operator as $\texttt{PoseGenerating}(\boldsymbol{c}, r, \{\theta\}) \rightarrow \{ \boldsymbol{p}_{ee} \}$. Given distances for approaching poses and cutting poses, i.e., $r^{a} \geq r^{c}$, we can generate potential approaching pose set $\{ \boldsymbol{p}^{a}_{ee}\}$ and cutting pose set $\{ \boldsymbol{p}^{c}_{ee} \}$ by feeding a set of angles $\{\theta\}$ where $ \theta \in [-\pi, \pi]$ to the operator. 

\subsubsection{Inverse Kinematics Set}

Naturally, after deciding on an end-effector pose, the planner needs to calculate the corresponding IK solution, i.e., joints configuration, with which the robot can achieve the desired pose, i.e., $\texttt{IK}(\boldsymbol{p}_{ee}) \rightarrow \boldsymbol{q}$, where $\boldsymbol{q}$ is the joint configuration. However, normal IK solvers typically provide a single IK solution. Such a single solution can lead to planning failures because (1) it's highly possible that the configuration of the robot will collide with intrinsic tree obstacles and (2) the configuration of the approaching pose needs to account for the following approach stage's motion, ensuring that the motion does not exceed the joint limits. In order to tackle the problems, we employ a learning-based diverse IK solver, i.e., IKFlow \cite{9793576}. IKFlow is a novel IK solver that can generate hundreds to thousands of diverse IK solutions rapidly once trained, therefore we can easily get a set of IK solutions for one end-effector pose. The distribution of the solution can be controlled by the latent distribution provided by the user. We utilize this IK solution set in two ways. First, compared to a single IK solution, the IK solution set enriches the solution space of the holistic planning problem. Second, it allows the planner to choose a robot configuration that satisfies constraints such as avoiding collisions and maintaining ample margin from the joint limits, enabling the subsequent motion to be conducted without exceeding the joint limits or encountering singularities. We formulate this step as $\texttt{IKFlow}(\boldsymbol{p}_{ee}) \rightarrow \{ \boldsymbol{q} \}$. 

\subsubsection{Holistic Pose-Motion Planning}


The two-stage motion can be formulated as a hierarchical planning problem, which is normally solved by a downstream planning strategy. For example, the planner first selects an end-effector pose. Then, an IK solver provides a single IK solution for the chosen pose. Finally, a motion planner such as Rapidly-exploring Random Tree (RRT) or its variants can be used to find a path in the joint space that avoids all the obstacles. However, such a downstream planning strategy will be difficult to advance in our case because the two stages of motion have different requirements: The moving stage avoids collisions without constraints in the end-effector's Cartesian space, while the approaching stage requires the end-effector moving in a bounded Cartesian space. 
Additionally, the configuration of the approaching pose should consider not only the accessibility of the moving stage but also the following approaching stage, to avoid the following motion reaching joint limits or singularities. To address the challenge, we take advantage of the redundancies in both Cartesian space and joint space. On top of that, we defined a cost function that evaluates the approaching poses and the configurations in their corresponding IK solution set. Finally, it selects the optimal configuration $\boldsymbol{q^{*}}$, which implies the optimal poses $\boldsymbol{p}_{ee}^{a, *}$ and $\boldsymbol{p}_{ee}^{c, *}$. We define the cost function $\mathcal{L}$ as follows:
\begin{equation}
    \mathcal{L}(\boldsymbol{q}, \mathcal{O}_{tree}) = \mathcal{H}(\boldsymbol{q}, \mathcal{O}_{tree}) + \mathcal{J}(\boldsymbol{q}) 
\end{equation}
\begin{equation}
    \mathcal{H} (\boldsymbol{q}, \mathcal{O}_{tree})= 
    \begin{cases}
    \infty & \textbf{if}\:\texttt{Collision}(\boldsymbol{q}, \mathcal{O}_{tree}) = \texttt{True} \\
    0 & \textbf{otherwise}
    \end{cases} 
\end{equation}

\begin{equation}
\mathcal{J}(\boldsymbol{q})=\frac{1}{m(\boldsymbol{q})}
=\frac{1}{\sqrt{\operatorname{det}\left(\hat{\mathbf{J}}(\boldsymbol{q}) \hat{\mathbf{J}}(\boldsymbol{q})^{\top}\right)}}
\end{equation}

Where $\mathcal{H}$ is the collision cost. By giving infinite cost when colliding, we can avoid the colliding configurations. We use the reciprocal of the Yoshikawa manipulability index $m$ for the joint limit cost $\mathcal{J}$, where $\hat{\mathbf{J}}(\boldsymbol{q}) \in \mathbb{R}^{3 \times n}$ is the translational rows of the robot's Jacobian.
The rationale behind this design is as follows: First, $m$ is a non-negative scalar that quantifies how easily the manipulator can achieve arbitrary velocities \cite{haviland2023manipulator}. Second, as the joint configuration $\boldsymbol{q}$ approaches a singularity, $m$ approaches zero, causing the cost to increase dramatically toward infinity \cite{yoshikawa1985manipulability}. 
This cost function encourages the planner to choose a configuration with sufficient ability to execute the following motions. Finally, since the intricate collision model will significantly increase the planning time with traditional motion planning methods, we deploy an advanced vector-accelerated motion planning (VAMP) method introduced in \cite{capt_2024}. The pseudo-code of the holistic pose-motion planning algorithm is shown in Algorithm \ref{alg:plan_alg}.


\begin{algorithm}
\caption{Holistic Pose-Motion Planning}\label{alg:plan_alg}

\textbf{Given} a cutting location set $\mathcal{C} = \{\boldsymbol{c}^{i}\}$, the obstacle model of the tree $\mathcal{O}_{tree}$, $r_a$, $r_c$ and $\{\theta\}$. \\

\For{every $\boldsymbol{c}^{i}$}{
    \textcolor{blue}{\# Generate approaching poses and cutting poses}\\

    $\{\boldsymbol{p}_{ee}^{a}\} = \texttt{PoseGenerating}(\boldsymbol{c}^{i}_c, r_a, \{\theta\})$\\
    
    $\{\boldsymbol{p}_{ee}^{c}\} = \texttt{PoseGenerating}(\boldsymbol{c}^{i}_c, r_c, \{\theta\})$\\
    
    \For{every $\boldsymbol{p}_{ee, \theta}^{a}$ in $\{\boldsymbol{p}_{ee}^{a}\}$}{
        \textcolor{blue}{\# Pose-motion planning}\\
        
        $\{\boldsymbol{q}\}_{\theta} = \texttt{IKFlow}(\boldsymbol{p}_{ee}^{a})$\\

        \If{$\{\boldsymbol{q}\}_{\theta} = \emptyset$}{\textbf{Continue}}
        \Else{
            \For{every $\boldsymbol{q}$ in $\{\boldsymbol{q}\}_{\theta}$}{
                $\mathcal{L}(\boldsymbol{q}) = \mathcal{H}(\cdot) + \mathcal{J}(\cdot)$\\
            }
        }
    }
    
    $\boldsymbol{q}^{*}, \theta^{*} = \underset{\boldsymbol{q} \in \{\boldsymbol{q}\}, \theta \in \{ \theta \}}{\arg\min}\, \mathcal{L}(\boldsymbol{q})$ \\

    $\tau_a = \texttt{VAMP}(\boldsymbol{q_0}, \boldsymbol{q}^{*})$ \\

    \textcolor{blue}{\# Get the cutting pose with the same angle}\\
    $\boldsymbol{p}_{ee}^{c, *} = \boldsymbol{p}_{ee, \theta}^{c}$ \\
}
\end{algorithm}


\subsubsection{Approaching Control}
After the end-effector arrives at the approaching pose, where the cutting position is close, the next step is to reach the actual cutting pose where the cutting position on the branch will be located between the blades of the shear. This step requires high accuracy and robust execution since misplacing the shear will cause failure or damage.
Therefore, a closed-loop controller is deployed in this stage. We use position-based servoing (PBS) \cite{haviland2023manipulator}. The advantage of PBS is that it can instantaneously change the end-effector velocity, this feature allows the controller to adjust the output on the fly according to the sensory information. PBS can be formulated as:
\begin{equation}\label{equ:pbs}
\begin{split}
    \dot{\boldsymbol{q}} & = \mathbf{J}(\boldsymbol{q})^{+} \cdot \boldsymbol{\nu} \\
    \boldsymbol{\nu} & = \boldsymbol{k}\boldsymbol{e}_{ee} \\
    \boldsymbol{e}_{ee} & = (\boldsymbol{p}^{c, *}_{ee} - \boldsymbol{p}_{ee}) \in \mathbb{R}^6
\end{split}
\end{equation}
Where $\mathbf{J}(\boldsymbol{q})$ is manipulator Jacobian, $\boldsymbol{\nu} = (\boldsymbol{\upsilon}, \boldsymbol{\omega})$ is the end-effector velocity, and $\boldsymbol{k} = \text{diag}(k_t, k_t, k_t, k_r, k_r, k_r)$ is a proportional gain, we use $k_t$ and $k_r$ for translation and rotation respectively because of their different units. $\boldsymbol{e}_{ee}$ is the end-effector pose error between the goal cutting pose $\boldsymbol{p}^{c, *}_{ee}$ and the current pose $\boldsymbol{p}_{ee}$. PBS allows the use of alternative sensor inputs, such as images.


\section{Experiments}\label{sec:experiments}

In this section, we present experiments to evaluate the performance of our method. First, we introduce a tree dataset collected from a real orchard. Next, we run repeated experiments on the dataset to compare our planning method with baseline approaches. Finally, we implement the workflow in a lab environment using a physical manipulator and real apple tree samples.

\subsection{Tree Dataset}\label{sec:tree_dataset}
\begin{figure}
    \centering
    \vspace{2 mm}
    \subfloat[]{\includegraphics[width=0.475\textwidth]{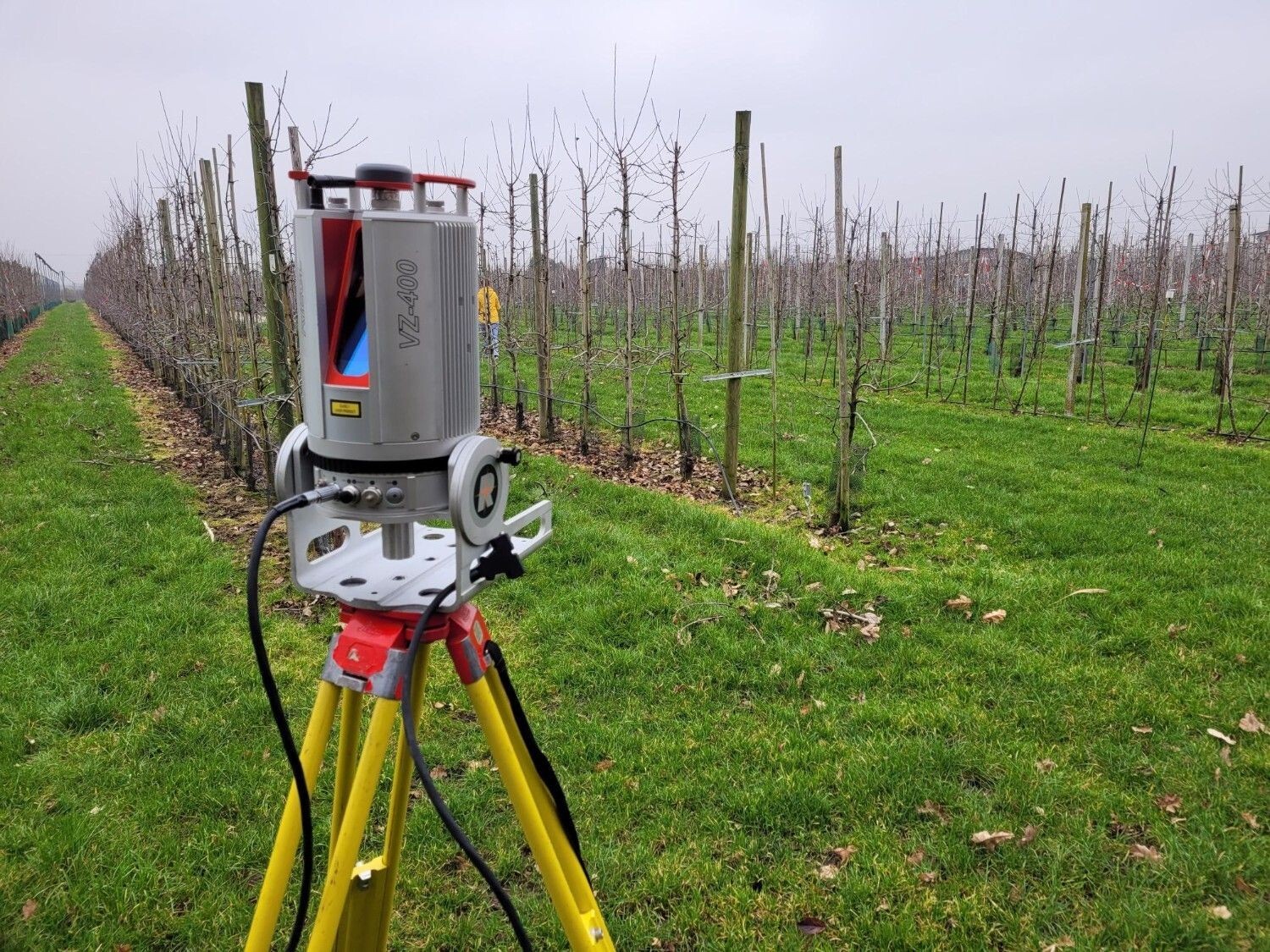}\label{fig:reiglscanner}}
    \hfil
    \subfloat[]{\includegraphics[width=0.236\textwidth]{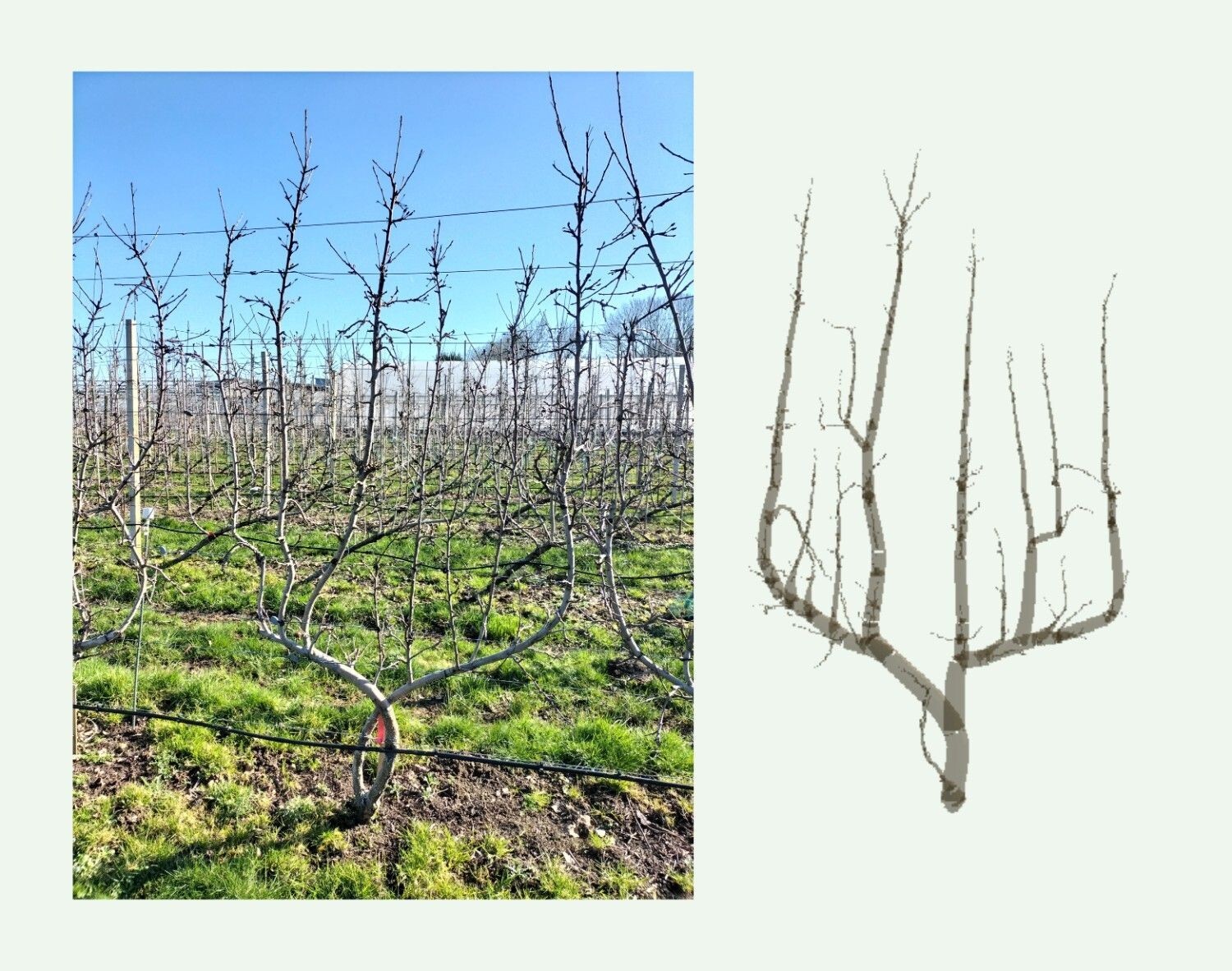}\label{fig:2d_tree}}
    \hfil
    \subfloat[]{\includegraphics[width=0.236\textwidth]{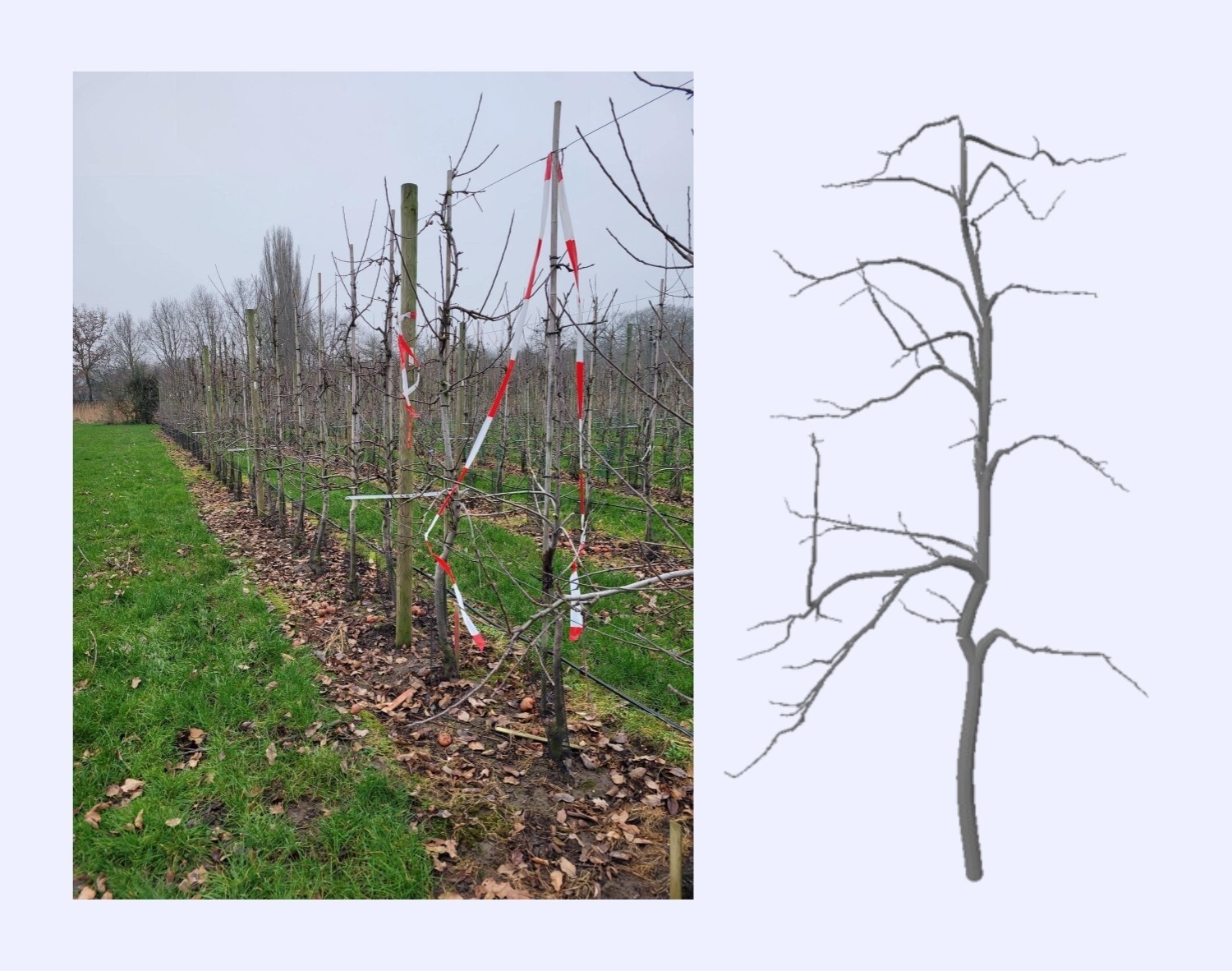}\label{fig:spindle_tree}}

    \caption{(a) This image shows the RIEGL VZ-400 laser scanner, which is used to perform the scans of the orchard. Note that in the lab, the Azure Kinect camera is currently used to obtain the point cloud. However, these point clouds are less dense, and it is challenging to fit the 2.5-meter tree fully into the camera's field of view. Both (b) and (c) show trees scanned in the WUR Randwijk orchard, along with a typical mesh representation of the architecture. (b) An example of a tree with a 2D structure. (c) An example of trees with a spindle structure.}
    \label{fig:scanner}
\end{figure}

\begin{figure*}
    \centering
    \vspace{2 mm}
    \includegraphics[width=0.98\textwidth]{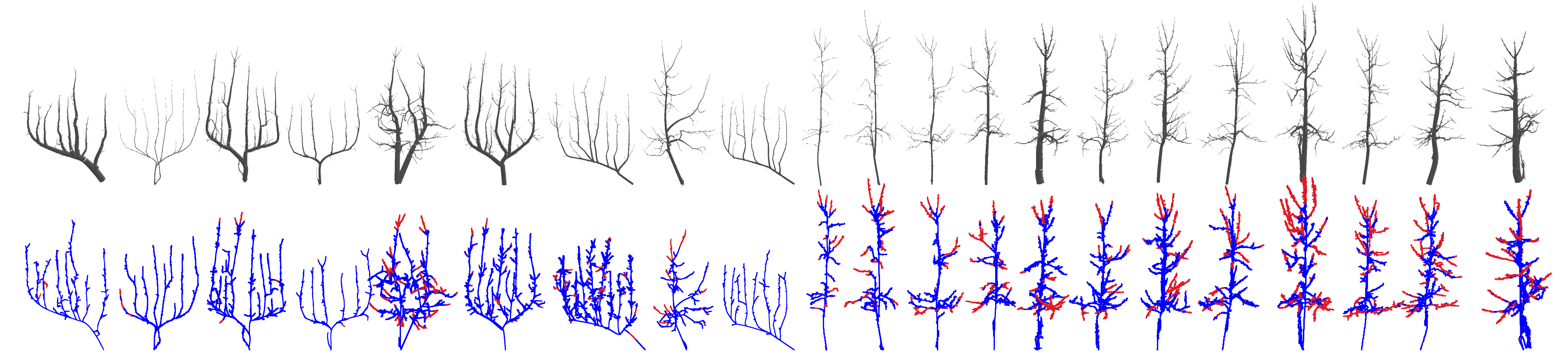}
    \caption{Real tree model dataset: Using the modeling strategy outlined in Section \ref{sec:tmacld}, we generate a dataset reconstructed from real trees. These trees are sampled from an actual orchard and vary in size and structure. The lower row shows the point clouds with cutting branch annotations (where red indicates the parts to be cut), and the upper row displays the AdTree models. We test our pruning algorithm on this dataset.}
    \label{fig:dataset}
\end{figure*}

To obtain highly accurate apple tree models for the dataset, Light Detection and Ranging (LiDAR) scans are created in the experimental orchard of Wageningen University (Proeftuin Randwijk in the Netherlands). The LiDAR scans are captured using the RIEGL VZ-400 laser scanner, shown in Figure \ref{fig:reiglscanner}.
There are two types of tree structures: 2D structure (Figure \ref{fig:2d_tree}) and spindle structure (Figure \ref{fig:spindle_tree}). The 2D structure makes apple trees more accessible to both robots and sunlight, while the spindle structure is the traditional apple architecture widely used in the Netherlands. In the orchard, we collect data from three rows of 2D apple trees, including both Elshof and Gaia varieties, and one row of spindle Elshof apple trees. We deliberately select apple trees with different growing techniques, and thus different structures, to thoroughly evaluate our pruning method. Notice that the pruning takes place in the winter, so the trees do not have leaves.
The point clouds are first manually split per tree using VR annotation software similar to \cite{garrido2021point}. Then, the point clouds are further annotated to identify the different objects in the orchard, namely the tree, wires, poles, drip-line, and the ground. This pre-processing step allows us to extract a point cloud containing only points that belong to the trees.
The dataset also includes pruning locations specified by the human operators. To collect the annotation label $\mathcal{L}_{\bullet}$ as used in Section \ref{sec:tmacld} for the cutting decisions, we scan the apple trees before (January $26^{th}$, 2022) and after (March $1^{st}$, 2022) pruning done by human. This provides us with information on which branches have been removed by performing change detection between the point clouds before and after pruning.
For change detection, the point clouds of the tree before and after pruning are first aligned using Iterative Closest Point (ICP). For each point before pruning, a \textit{kdtree} of the aligned point cloud after pruning identifies the points within a 5 cm distance. If no points are found, the points likely belong to a removed branch. This allows us to determine the labels $\mathcal{L}_{\bullet}$ per point for branches that have been removed. The results are shown in Figure \ref{fig:dataset}.

\subsection{Baseline Comparison}\label{sec:baseline_comparison}
\begin{figure}[h]
    \centering
    \subfloat[]{\includegraphics[width=0.19\textwidth]{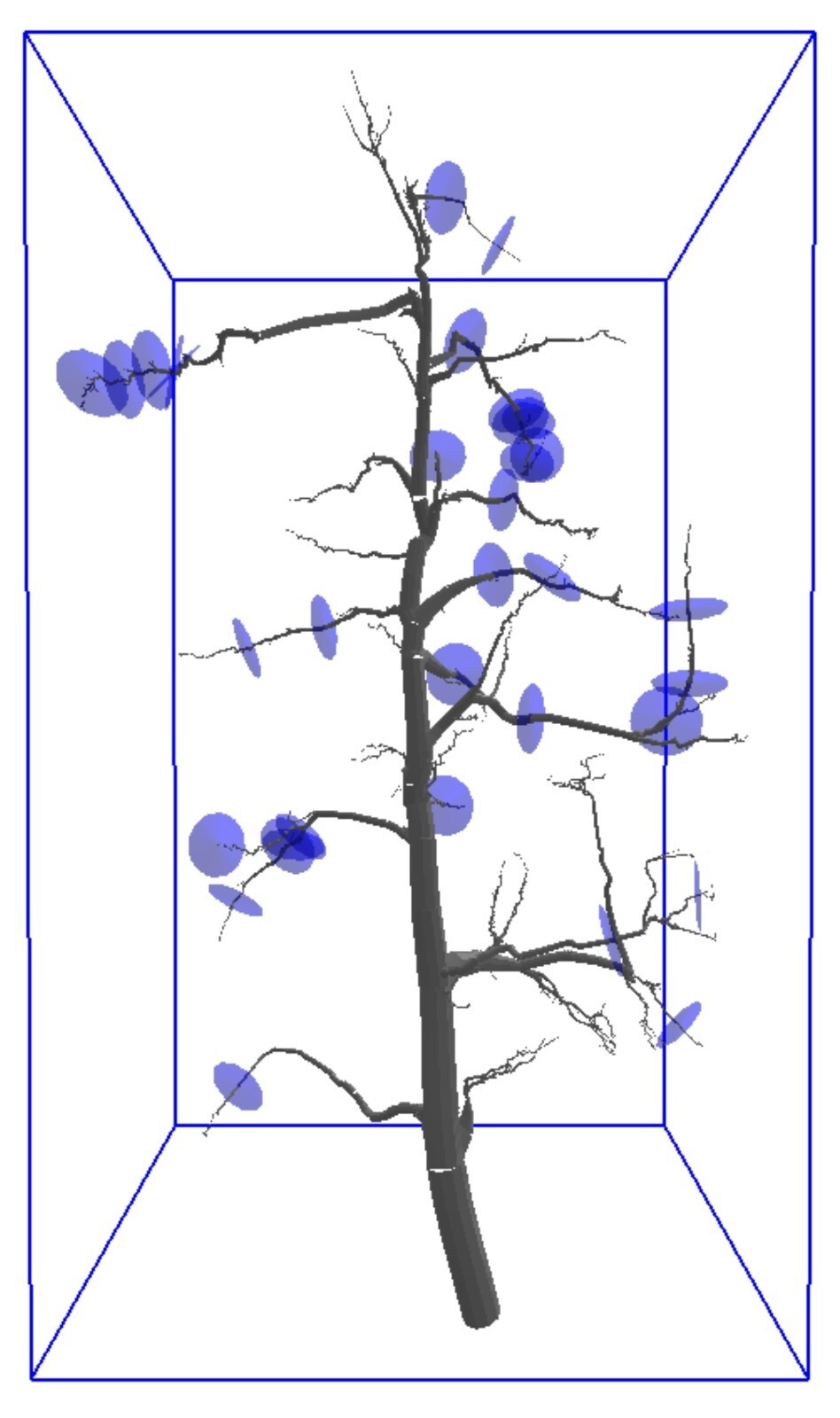}\label{fig:robot_poses_1}}
    \hfil
    \subfloat[]{\includegraphics[width=0.22\textwidth]{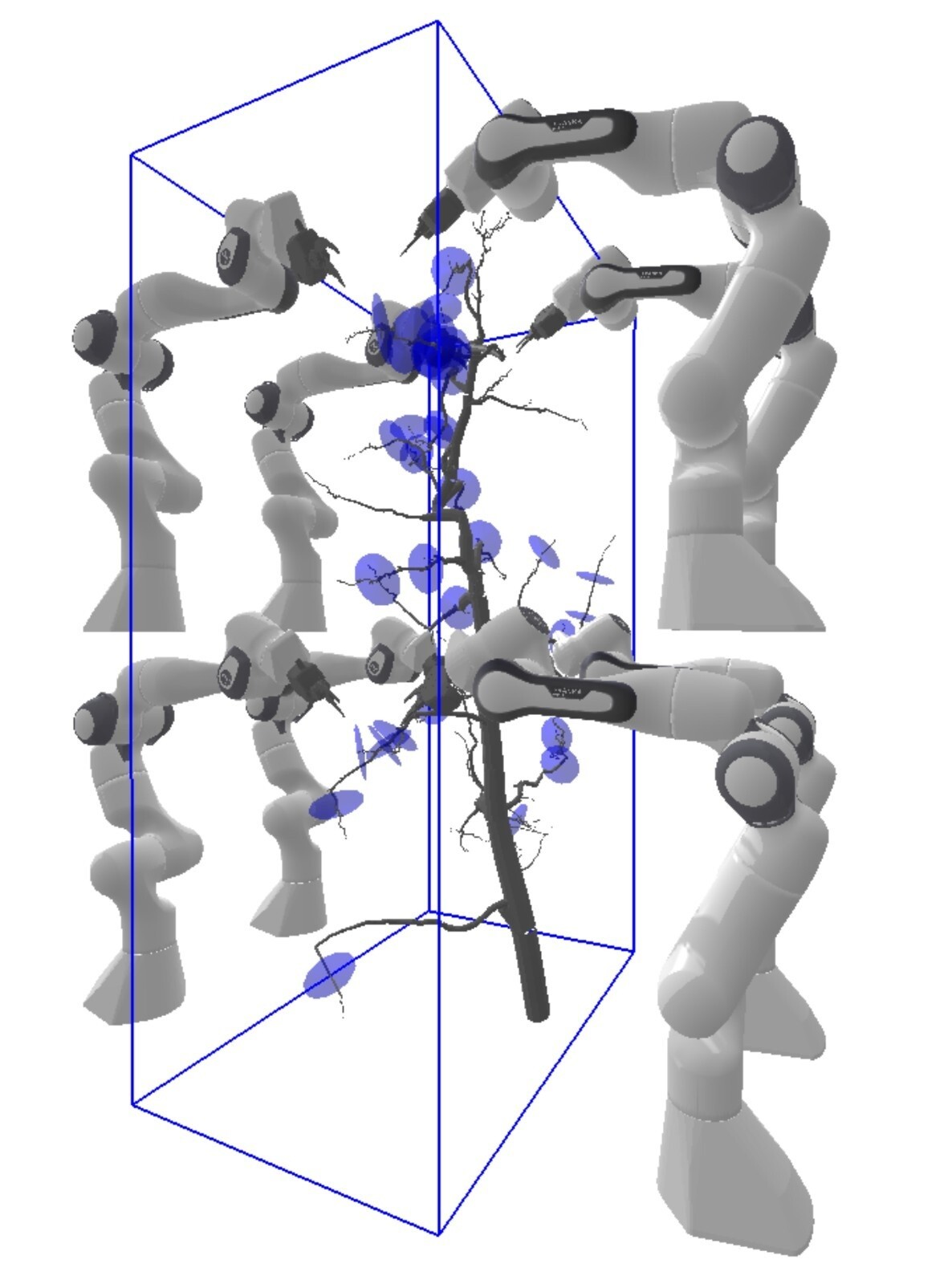}\label{fig:robot_poses_2}} 
    \caption{Comparison experiment setting. (a) We first calculate the axis-aligned bounding box of the model. (b) We assume the robot's base can be steered to 8 relative poses of the tree. The relative pose is defined by the size of the bounding box because of the scale difference between different trees. }
    \label{fig:robot_poses}
\end{figure}

\begin{figure}[h]
    \centering
    \includegraphics[width=0.48\textwidth]{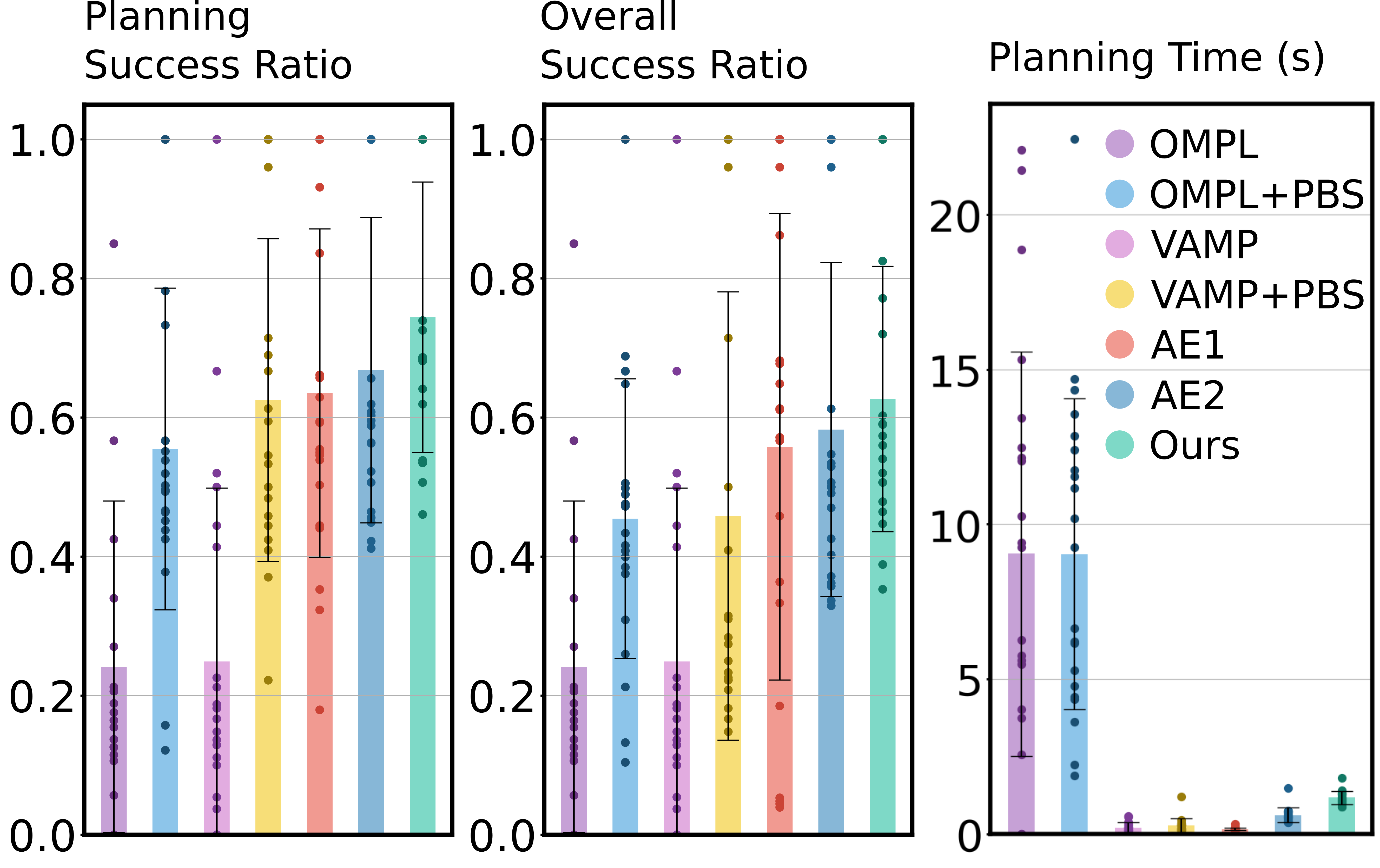}
    \caption{Comparison experiment results. We conduct experiments on all samples in the dataset and calculate the success ratios for each tree. From left to right are the planning success ratio, overall success (successful planning and execution) ratio, and planning time. 
    The bars indicate the average value, the dark lines on top of the bars are the standard deviation, the dots illustrate the raw values. It can be seen that our proposed method outperforms the baseline models in the success ratios. Regarding the planning time, the VAMP-based methods in general achieved much faster planning speed. }
    \label{fig:exp_bl_results}
\end{figure}
In this section, we implement two benchmarking motion planning libraries, Open Motion Planning Library (OMPL) and Vector-Accelerated Motion Planning (VAMP), in the same pruning scenarios. The six baseline methods can be categorized into three subclasses:
\begin{itemize}
\item Elementary motion planning: The first and third baselines use RRT-Connect (RRTC), as proposed in \cite{chen2022path}, implemented with OMPL and VAMP, respectively.

\item Two-stage motion planning: The second and fourth baselines follow a two-stage approach similar to our proposed method, where RRTC is plainly combined with PBS. The key difference lies in the RRTC implementation: one uses OMPL (OMPL + PBS), while the other uses VAMP (VAMP + PBS).

\item Ablation experiments: The fifth baseline (AE1) is a variation of our method that removes redundancy in the Cartesian space. Specifically, it eliminates the pose-generating module from the workflow. The sixth baseline (AE2) retains Cartesian space redundancy but removes joint space redundancy. In this case, we replace IKFlow with a conventional single-solution inverse kinematics (IK) solver.

\end{itemize}
Visualization and planning are conducted in PyBullet, a widely used robotic simulation environment \cite{coumans2016pybullet}.
The substitution IK solver used in baseline methods is TRAC-IK \cite{7363472}, which is a widely used numerical IK solution. However, it can only output one IK solution per call, and no diverse solutions are guaranteed. We assume that the base of the robot can be lifted freely in a constrained 2D space. To obtain more experiment trials, we set the robot base at eight different locations (shown in Figure \ref{fig:robot_poses_2}) for each tree. The locations are defined with the axis-aligned bounding box of the tree model. We implement baselines and our method on all 21 tree samples in the data set introduced in Section \ref{sec:tree_dataset}. The methods are compared with three key metrics: Planning success ratio, overall success ratio, and planning time for successful instances. We identify planning failures when no collision-free path or configurations are found before the timeout occurs. The timeout for planning is set at 30 seconds. Achieving overall success requires both successful planning and execution. The execution failures are identified when (1) any joint angle in $\boldsymbol{q}$ reaches the limits; (2) any joint velocity command in $\dot{\boldsymbol{q}}$ reaches the velocity limits and (3) any collision is detected. The singularity failures fall in the joint velocity failure because when a robot reaches a singularity, the Jacobian's inverse $\mathbf{J}(\boldsymbol{q})^{+}$ in Equation (\ref{equ:pbs}), therefore the velocity command $\nu$, will result in extremely large values.

As shown in Figure \ref{fig:exp_bl_results}, our method achieves a higher success rate compared to the baselines. In general, the elementary methods struggle to achieve high success rates, while the two-stage methods show significant improvements. This is because directly planning to the cutting configuration requires navigating through narrow free spaces among branch obstacles. The two ablation experiments highlight the benefits of incorporating the pose-generating and IKFlow modules, respectively.
Regarding planning time, OMPL is generally slower than VAMP due to VAMP’s vector-acceleration features. Our method takes slightly longer to find a solution compared to the plain VAMP implementation and ablation methods. This additional computation time is primarily due to the holistic pose-motion planning process, which requires extra calculations for manipulability and collision checking with each redundant configuration. As a result, the planning time increases as the inverse kinematics set expands.


\subsection{Real-world Experiments}

\begin{figure}
    \centering
    \subfloat[]{\includegraphics[width=0.2\textwidth]{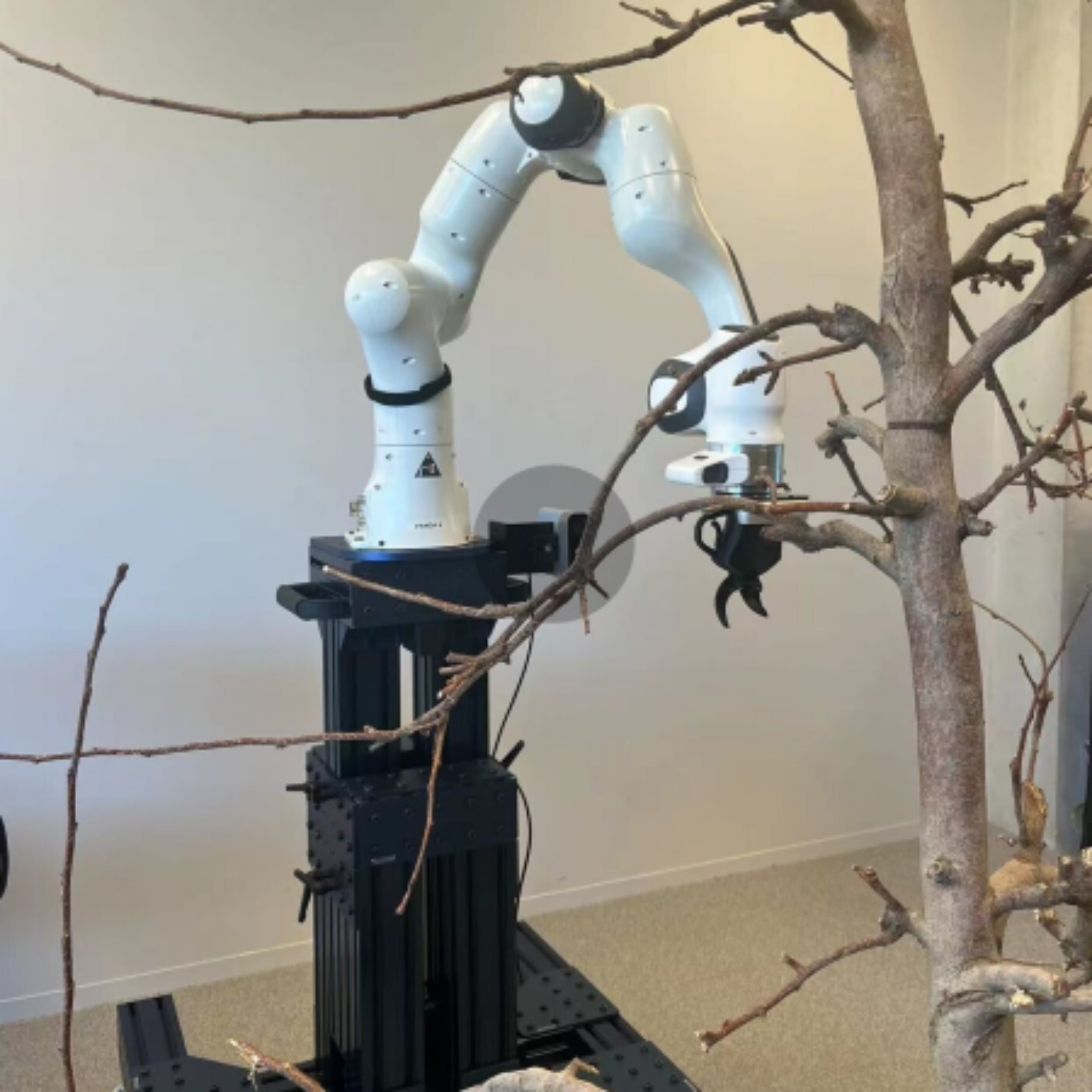}\label{fig:real_1}}
    \hfil
    \subfloat[]{\includegraphics[width=0.2\textwidth]{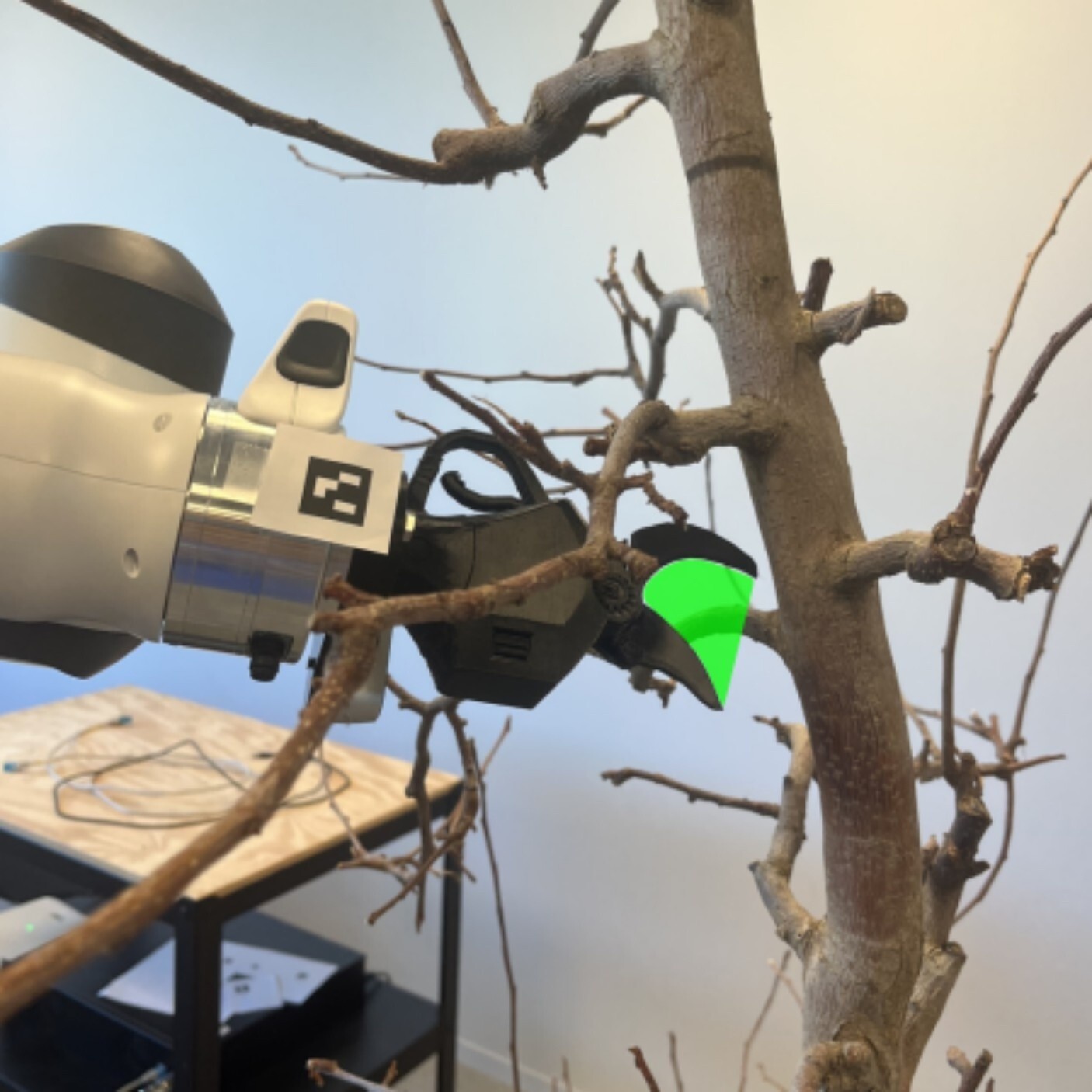}\label{fig:real_2}} 
    
    \caption{The real experiment setting. (a) The setting is compressed with a robot mount, an Azure Kinect depth camera (marked in the shade), and a Franka Emika manipulator. The tree sample has a complex spindle structure. (b) The manipulator's end-effector is replaced with a pruner. A successful execution occurs when the branch is cut at a point within the green area. }
    \label{fig:real_exp_setup}
\end{figure}

\begin{figure}
    \centering
    \includegraphics[width=0.4\textwidth]{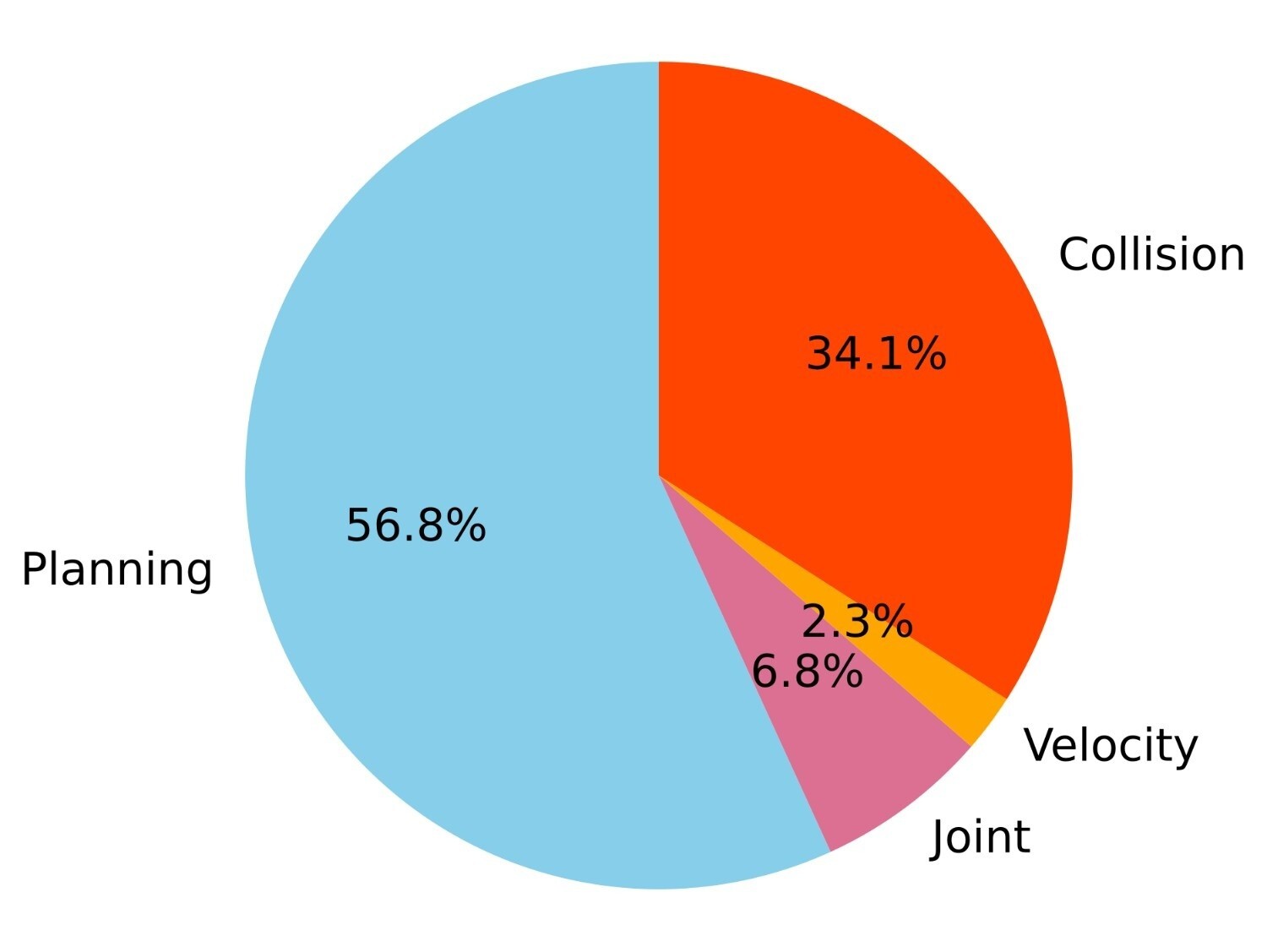}
    \caption{Diagnosis of failures. We analyze the causes of overall failures. The types of failures are detailed in Section \ref{sec:baseline_comparison}. During the real-world experiment, 44 overall failures occurred. These include 25 planning failures (in blue), 3 instances of exceeding joint limits (in pink), 1 instance of exceeding joint velocity limits (in yellow), and 15 collisions (in red).}
    \label{fig:diagnose}
\end{figure}

Finally, we implemented the proposed system in a real-world setting using a physical robotic manipulator. As shown in Figure \ref{fig:real_exp_setup}, the hardware system includes a stand on which the robot is mounted, an Azure Kinect camera, and a Franka Emika manipulator. We replaced the default gripper with a 3D-printed model based on a market-available electric pruner. The height of the stand can be adjusted freely within a certain limit, allowing for similar robot base poses as in the simulation experiments. The tree sample is a real tree disposed of from the experimental orchard. Given the limited number of real tree samples, we rotated the tree to create different collision scenarios. In each trial, we use different relative positions of the robot and the tree, similar to the approach used in the experiments described in Section \ref{sec:baseline_comparison}. The Azure Kinect ToF camera is used to obtain a point cloud of the tree. 
In this case, a couple of simple filters, such as Euclidean clustering and axis-aligned bounding box, are applied to ignore points that do not belong to the tree. The cutting locations are determined by human experts who annotate the point clouds using a virtual reality annotation tool to generate point labels $\mathcal{L}_{\bullet}$. The method is similar to the one described in \cite{garrido2021point}, conducted with an Oculus Quest 2 headset. 
After which we can proceed with the automatic pipeline described in the paper. It is worth noting that collision detection in real-world experiments is realized by checking the external wrench.
The experiments show that the planning success ratio can be kept at 0.73 (69 success out of 94 cuts). The overall success ratio, however, drops to 0.53 (50 success out of 94 cuts). We analyzed the cause of the failures during the implementation of the workflow, the result is shown in Figure \ref{fig:diagnose}. Each section shows the percentage of each failure type contributing to the total number of failures. The result shows that besides planning failure, most failures are caused by collisions during the execution. We observed that such collisions are caused by AdTree modeling imperfections: First, the AdTree model might miss some small branches, or branches occluded by others, that are not considered in the planning phase; The short offshoots near a cutting location can cause collision in the approaching stage of motions.
However, neglecting the collision in the approaching stage is crucial because, as demonstrated in the first baseline, directly planning motion to the cutting pose results in a low success ratio. This issue can be addressed by local observation with an in-hand camera that we leave for future work. 

\subsection{Limitations}
As we focus on increasing the capability of the robotic pruning task, 
the intuition behind our method is straightforward: extending the search space for such a planning problem with a narrow solution space. However, our method requires more solving time compared to the control-based baseline, which considers only a single approach pose and configuration, and no motion planning procedure is included. Another limitation we discovered during experiments is the trade-off between collision avoidance and completeness of the pruning tasks, that is, the planning tends to fail when the collision model becomes over-detailed. To tackle this issue, the method needs to allow collision between the robot and some pliable branches. Thirdly, we assume the tree is static, the dynamic of the tree with disturbances such as wind should be considered in future work.


\section{Conclusion}\label{sec:conclusion}

In this paper, we addressed the planning problem associated with pruning tasks utilizing redundant robotic manipulators. The problem has been effectively formulated as a holistic planning problem, featuring multiple planning levels. We designed a comprehensive workflow from scanning and modeling to a holistic planning method leveraging the intrinsic redundancies of the pruning task. The experiments highlight the importance of incorporating holistic behavior planning in achieving more successful pruning operations. As we focus on the behavior of the robot, the proposed workflow requires human experts to decide on the cutting location.
Future research can be conducted on designing a complete system that encompasses all levels of decision-making to fully realize the potential of robotic pruning.



\medskip

\bibliographystyle{ieeetr}
\bibliography{reference}

\begin{thebibliography}{10}

\bibitem{tinoco2021review}
V.~Tinoco, M.~F. Silva, F.~N. Santos, L.~F. Rocha, S.~Magalh{\~a}es, and L.~C. Santos, ``A review of pruning and harvesting manipulators,'' in {\em 2021 IEEE International Conference on Autonomous Robot Systems and Competitions (ICARSC)}, pp.~155--160, IEEE, 2021.

\bibitem{ferree1993apple}
D.~C. Ferree and W.~T. Rhodus, ``Apple tree performance with mechanical hedging or root pruning in intensive orchards,'' {\em Journal of the American Society for Horticultural Science}, vol.~118, no.~6, pp.~707--713, 1993.

\bibitem{9811628}
A.~You, H.~Kolano, N.~Parayil, C.~Grimm, and J.~R. Davidson, ``Precision fruit tree pruning using a learned hybrid vision/interaction controller,'' in {\em 2022 International Conference on Robotics and Automation (ICRA)}, pp.~2280--2286, 2022.

\bibitem{9197551}
A.~You, F.~Sukkar, R.~Fitch, M.~Karkee, and J.~R. Davidson, ``An efficient planning and control framework for pruning fruit trees,'' in {\em 2020 IEEE International Conference on Robotics and Automation (ICRA)}, pp.~3930--3936, 2020.

\bibitem{you2024realtimehardwareagnosticframework}
A.~You, A.~Mehta, L.~Strohbehn, J.~Hemming, C.~Grimm, and J.~R. Davidson, ``A real-time, hardware agnostic framework for close-up branch reconstruction using rgb data,'' 2024.

\bibitem{botterill2017robot}
T.~Botterill, S.~Paulin, R.~Green, S.~Williams, J.~Lin, V.~Saxton, S.~Mills, X.~Chen, and S.~Corbett-Davies, ``A robot system for pruning grape vines,'' {\em Journal of Field Robotics}, vol.~34, no.~6, pp.~1100--1122, 2017.

\bibitem{you2023semiautonomous}
A.~You, N.~Parayil, J.~G. Krishna, U.~Bhattarai, R.~Sapkota, D.~Ahmed, M.~Whiting, M.~Karkee, C.~M. Grimm, and J.~R. Davidson, ``Semiautonomous precision pruning of upright fruiting offshoot orchard systems: An integrated approach,'' {\em IEEE Robotics \& Automation Magazine}, 2023.

\bibitem{zahid2020collision}
A.~Zahid, L.~He, D.~D. Choi, J.~Schupp, and P.~Heinemann, ``Collision free path planning of a robotic manipulator for pruning apple trees,'' in {\em 2020 ASABE annual international virtual meeting}, p.~1, American Society of Agricultural and Biological Engineers, 2020.

\bibitem{rs11182074}
S.~Du, R.~Lindenbergh, H.~Ledoux, J.~Stoter, and L.~Nan, ``Adtree: Accurate, detailed, and automatic modelling of laser-scanned trees,'' {\em Remote Sensing}, vol.~11, no.~18, 2019.

\bibitem{fan2020adqsm}
G.~Fan, L.~Nan, Y.~Dong, X.~Su, and F.~Chen, ``Adqsm: A new method for estimating above-ground biomass from tls point clouds,'' {\em Remote Sensing}, vol.~12, no.~18, p.~3089, 2020.

\bibitem{keerthinathan2025modelling}
P.~Keerthinathan, M.~Winsen, T.~Krishnakumar, A.~Ariyanayagam, G.~Hamilton, and F.~Gonzalez, ``Modelling lidar-based vegetation geometry for computational fluid dynamics heat transfer models,'' {\em Remote Sensing}, vol.~17, no.~3, p.~552, 2025.

\bibitem{gao2025extraction}
J.~Gao, L.~Tang, H.~Su, J.~Chen, and Y.~Yuan, ``Extraction of tree branch skeletons from terrestrial lidar point clouds,'' {\em Ecological Informatics}, vol.~85, p.~102960, 2025.

\bibitem{you2022optical}
A.~You, C.~Grimm, and J.~R. Davidson, ``Optical flow-based branch segmentation for complex orchard environments,'' in {\em 2022 IEEE/RSJ International Conference on Intelligent Robots and Systems (IROS)}, pp.~9180--9186, IEEE, 2022.

\bibitem{borrenpohl2023automated}
D.~Borrenpohl and M.~Karkee, ``Automated pruning decisions in dormant sweet cherry canopies using instance segmentation,'' {\em Computers and Electronics in Agriculture}, vol.~207, p.~107716, 2023.

\bibitem{10384649}
S.~Häring, S.~Folawiyo, M.~Podguzova, S.~krauß, and D.~Stricker, ``Vid2cuts: A framework for enabling ai-guided grapevine pruning,'' {\em IEEE Access}, vol.~12, pp.~5814--5836, 2024.

\bibitem{LI2023108149}
Y.~Li, Z.~Zhang, X.~Wang, W.~Fu, and J.~Li, ``Automatic reconstruction and modeling of dormant jujube trees using three-view image constraints for intelligent pruning applications,'' {\em Computers and Electronics in Agriculture}, vol.~212, p.~108149, 2023.

\bibitem{chen2022path}
Y.~Chen, Y.~Fu, B.~Zhang, W.~Fu, and C.~Shen, ``Path planning of the fruit tree pruning manipulator based on improved rrt-connect algorithm,'' {\em International Journal of Agricultural and Biological Engineering}, vol.~15, no.~2, pp.~177--188, 2022.

\bibitem{zahid2020development}
A.~Zahid, M.~S. Mahmud, L.~He, D.~Choi, P.~Heinemann, and J.~Schupp, ``Development of an integrated 3r end-effector with a cartesian manipulator for pruning apple trees,'' {\em Computers and Electronics in Agriculture}, vol.~179, p.~105837, 2020.

\bibitem{silwal2021bumblebee}
A.~Silwal, F.~Yandun, A.~Nellithimaru, T.~Bates, and G.~Kantor, ``Bumblebee: A path towards fully autonomous robotic vine pruning. arxiv 2021,'' {\em arXiv preprint arXiv:2112.00291}, 2021.

\bibitem{molaei2022kinematic}
F.~Molaei and S.~Ghatrehsamani, ``Kinematic-based multi-objective design optimization of a grapevine pruning robotic manipulator,'' {\em AgriEngineering}, vol.~4, no.~3, pp.~606--625, 2022.

\bibitem{9562075}
F.~Yandun, T.~Parhar, A.~Silwal, D.~Clifford, Z.~Yuan, G.~Levine, S.~Yaroshenko, and G.~Kantor, ``Reaching pruning locations in a vine using a deep reinforcement learning policy,'' in {\em 2021 IEEE International Conference on Robotics and Automation (ICRA)}, pp.~2400--2406, 2021.

\bibitem{YOU2022106622}
A.~You, C.~Grimm, A.~Silwal, and J.~R. Davidson, ``Semantics-guided skeletonization of upright fruiting offshoot trees for robotic pruning,'' {\em Computers and Electronics in Agriculture}, vol.~192, p.~106622, 2022.

\bibitem{9793576}
B.~Ames, J.~Morgan, and G.~Konidaris, ``Ikflow: Generating diverse inverse kinematics solutions,'' {\em IEEE Robotics and Automation Letters}, vol.~7, no.~3, pp.~7177--7184, 2022.

\bibitem{haviland2023manipulator}
J.~Haviland and P.~Corke, ``Manipulator differential kinematics: Part i: Kinematics, velocity, and applications,'' {\em IEEE Robotics \& Automation Magazine}, 2023.

\bibitem{yoshikawa1985manipulability}
T.~Yoshikawa, ``Manipulability of robotic mechanisms,'' {\em The international journal of Robotics Research}, vol.~4, no.~2, pp.~3--9, 1985.

\bibitem{capt_2024}
C.~W. Ramsey, Z.~Kingston, W.~Thomason, and L.~E. Kavraki, ``Collision-affording point trees: {SIMD}-amenable nearest neighbors for fast collision checking,'' in {\em Robotics: Science and Systems}.
\newblock To Appear.

\bibitem{garrido2021point}
D.~Garrido, R.~Rodrigues, A.~Augusto~Sousa, J.~Jacob, and D.~Castro~Silva, ``Point cloud interaction and manipulation in virtual reality,'' in {\em 2021 5th International Conference on Artificial Intelligence and Virtual Reality (AIVR)}, pp.~15--20, 2021.

\bibitem{coumans2016pybullet}
E.~Coumans and Y.~Bai, ``Pybullet, a python module for physics simulation for games, robotics and machine learning,'' 2016.

\bibitem{7363472}
P.~Beeson and B.~Ames, ``Trac-ik: An open-source library for improved solving of generic inverse kinematics,'' in {\em 2015 IEEE-RAS 15th International Conference on Humanoid Robots (Humanoids)}, pp.~928--935, 2015.

\end{thebibliography}

\end{document}